\documentclass[a4paper,conference]{IEEEtran}
\usepackage[pdftex]{graphicx}
\usepackage{comment}
\usepackage{amsmath,amssymb} % define this before the line numbering.
\usepackage{color}
\usepackage{subcaption}
\usepackage{booktabs}
\usepackage{varwidth}
\usepackage{makecell}
\usepackage{array}
\usepackage{epsfig}
\usepackage{lipsum}
\usepackage{verbatim}
\usepackage{dsfont}

\newcommand{\etal}{et~al\mbox{.}}
\definecolor{MTKGold}{RGB}{241, 154, 33}

\ifCLASSINFOpdf
\else
\fi
\begin{document}
\title{Explorable Tone Mapping Operators}

%\author{\IEEEauthorblockN{Chien Chuan Su}
%\IEEEauthorblockA{Graduate Institute of\\Communication Engineering\\
%National Taiwan University\\
%Taipei, Taiwan 106\\
%Email: ccs@xxx.xxx}
%\and
%\IEEEauthorblockN{Yu Lun Liu}
%\IEEEauthorblockA{MediaTek Inc.\\Hsinchu, Taiwan 300\\
%Email: Yu-Lun Liu@mediatek.com}}


\author{\IEEEauthorblockN{Chien-Chuan Su\IEEEauthorrefmark{1}\IEEEauthorrefmark{2},
Ren Wang\IEEEauthorrefmark{2},
Hung-Jin Lin\IEEEauthorrefmark{2},
Yu-Lun Liu\IEEEauthorrefmark{2},
Chia-Ping Chen\IEEEauthorrefmark{2},
Yu-Lin Chang\IEEEauthorrefmark{2} and
Soo-Chang Pei\IEEEauthorrefmark{1}}
\IEEEauthorblockA{\IEEEauthorrefmark{1}
National Taiwan University,
Taipei, Taiwan\\ Email: $\{$r06942145, peisc$\}$@ntu.edu.tw}
\IEEEauthorblockA{\IEEEauthorrefmark{2}MediaTek Inc., Hsinchu, Taiwan\\
Email: $\{$ren.wang, hungjin.lin, yu-lun.liu, chiaping.chen, yulin.chang$\}$@mediatek.com}}


\twocolumn[{%
\renewcommand\twocolumn[1][]{#1}%
\maketitle
\begin{center}
    \captionsetup{type=figure}
% 	\footnotesize
    \begin{minipage}[c]{1.0\textwidth}
    \centering
        \includegraphics[width=1.0\textwidth]{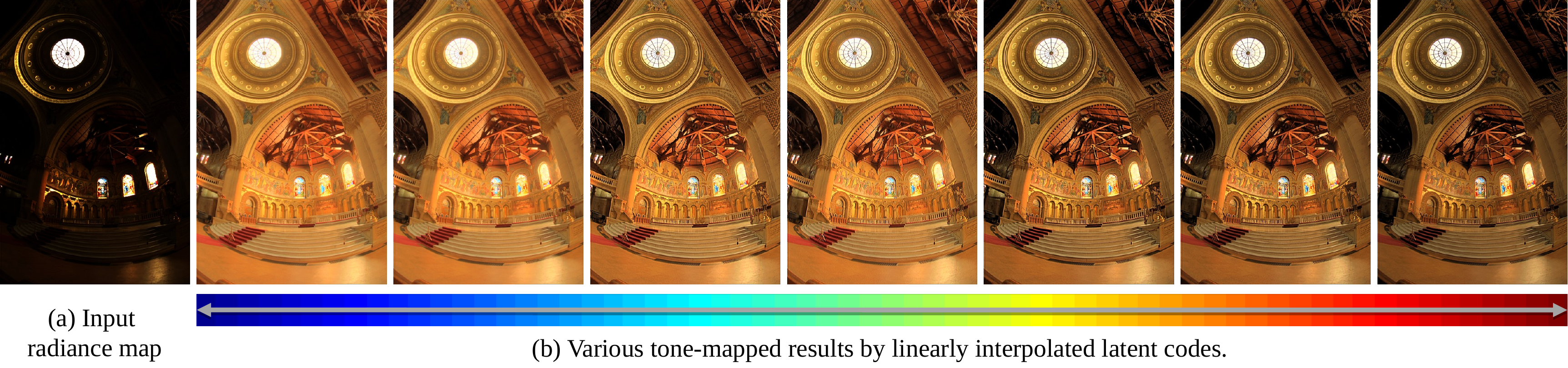}
    \end{minipage}
    \\
    \caption{
        \textbf{Various tone-mapped results from a single input with a single model.} Given an input HDR radiance map, our method can produce various reasonable tone-mapped LDR images by adjusting the latent code.
    } 
    \label{fig:teaser}
\end{center}
}]

%\begin{abstract}
%The abstract goes here.
%\end{abstract}
\begin{abstract}
% Note that the maximum length of abstract is 150 words.
Tone-mapping plays an essential role in high dynamic range (HDR) imaging.
It aims to preserve visual information of HDR images in a medium with a limited dynamic range.
Although many works have been proposed to provide tone-mapped results from HDR images, most of them can only perform tone-mapping in a single pre-designed way.
However, the subjectivity of tone-mapping quality varies from person to person, and the preference of tone-mapping style also differs from application to application.
In this paper, a learning-based multimodal tone-mapping method is proposed, which not only achieves excellent visual quality but also explores the style diversity.
%In this paper, a learning-based steerable tone-mapping method is proposed, which not only achieves excellent visual quality but also enables easy style adjustability.
%
Based on the framework of BicycleGAN~\cite{NIPS2017_6650}, the proposed method can provide a variety of expert-level tone-mapped results by manipulating different latent codes.
%Based on the framework of an improved cVAE-GAN~\cite{NIPS2017_6650}, the proposed method can provide a variety of expert-level tone-mapping preferences by manipulating different latent codes. 
%
% Moreover, the proposed method suffers from minimal artifacts among both learning based and non-learning based methods.
%
Finally, we show that the proposed method performs favorably against state-of-the-art tone-mapping algorithms both quantitatively and qualitatively.
%
%\keywords{Tone mapping, computational photography, generative adversarial network}
\end{abstract}


% \fi
\IEEEpeerreviewmaketitle

\section{Introduction}
\label{sec:intro}

%%%%%%%%%%%%%%%%%%%%%%%%%%%%%%%%%%%%%%%%%%%%%%%%%%%%%%%%%%%%%%%%%%%%%%%%%%%%%%%%%%%%
% what is HDR
%%%%%%%%%%%%%%%%%%%%%%%%%%%%%%%%%%%%%%%%%%%%%%%%%%%%%%%%%%%%%%%%%%%%%%%%%%%%%%%%%%%%
In the real world, the dynamic range (DR) of a natural scene is often too wide ($\text{DR} > 10^{7}$) for a camera to capture, especially for direct light sources such as the sun. Thanks to the development of the multi-exposure fusion technique~\cite{5936115}, we can fuse all the detail from images with different exposures into a single high dynamic range (HDR) image. 

%%%%%%%%%%%%%%%%%%%%%%%%%%%%%%%%%%%%%%%%%%%%%%%%%%%%%%%%%%%%%%%%%%%%%%%%%%%%%%%%%%%%
% what is tone mapping
%%%%%%%%%%%%%%%%%%%%%%%%%%%%%%%%%%%%%%%%%%%%%%%%%%%%%%%%%%%%%%%%%%%%%%%%%%%%%%%%%%%%
The HDR image contains rich visual information and needs a higher bit-depth to store the wide dynamic range data. Nevertheless, most display devices could only show low dynamic range images (LDR, often stored in 8-bit). Tone-mapping algorithms are then proposed to compress HDR images into LDR ones while trying to preserve the perceptual content as much as possible.
%

%%%%%%%%%%%%%%%%%%%%%%%%%%%%%%%%%%%%%%%%%%%%%%%%%%%%%%%%%%%%%%%%%%%%%%%%%%%%%%%%%%%%
% How tone mapping is done
%%%%%%%%%%%%%%%%%%%%%%%%%%%%%%%%%%%%%%%%%%%%%%%%%%%%%%%%%%%%%%%%%%%%%%%%%%%%%%%%%%%%
A series of tone-mapping algorithms have been proposed in the past two decades~\cite{durand_fast_2002,fattal_gradient_2002,reinhard_photographic_2002,liang_hybrid_2018}. 
Many of them decompose the HDR image into two parts: 
a \emph{base layer} that is often smoothed but still maintains the original global dynamic range,
and a \emph{detail layer} that possesses only local edge or detail information. 
Before fusing the base layer and the detail layer back into an LDR image, 
the base layer is usually compressed to reduce the dynamic range, 
while the detail layer is enhanced or boosted so that better visual content is preserved.

%%%%%%%%%%%%%%%%%%%%%%%%%%%%%%%%%%%%%%%%%%%%%%%%%%%%%%%%%%%%%%%%%%%%%%%%%%%%%%%%%%%%
% Decomposition matters 
%%%%%%%%%%%%%%%%%%%%%%%%%%%%%%%%%%%%%%%%%%%%%%%%%%%%%%%%%%%%%%%%%%%%%%%%%%%%%%%%%%%%
Within this scheme, low-frequency and high-frequency information in HDR images are handled separately so that the dynamic range can be significantly compressed while the local detail can still be preserved in LDR images.
Accordingly, the decomposition of an HDR image into a base layer and a detail layer greatly influences the quality of a tone-mapping method, and the way to perform the decomposition almost constitutes the main differences among different methods.

%%%%%%%%%%%%%%%%%%%%%%%%%%%%%%%%%%%%%%%%%%%%%%%%%%%%%%%%%%%%%%%%%%%%%%%%%%%%%%%%%%%%
% Detail enhancement leads to artifacts
%%%%%%%%%%%%%%%%%%%%%%%%%%%%%%%%%%%%%%%%%%%%%%%%%%%%%%%%%%%%%%%%%%%%%%%%%%%%%%%%%%%%
As for detail enhancement, some methods try to brighten areas around dark objects and result in halo artifacts~\cite{durand_fast_2002}. 
Some other methods over-emphasize the edge information, thus produce unrealistic over-enhanced results~\cite{Mantiuk:2006:PFC:1166087.1166095}. 
There are methods aiming to solve these problems, but they often work well only on specific types of images and need a lot of parameter tuning to obtain the best results~\cite{fattal_gradient_2002}. 
This tuning process is often time-consuming and hard to reproduce.

%%%%%%%%%%%%%%%%%%%%%%%%%%%%%%%%%%%%%%%%%%%%%%%%%%%%%%%%%%%%%%%%%%%%%%%%%%%%%%%%%%%%
% There are also deep learning based methods
%%%%%%%%%%%%%%%%%%%%%%%%%%%%%%%%%%%%%%%%%%%%%%%%%%%%%%%%%%%%%%%%%%%%%%%%%%%%%%%%%%%%
%To manage the time-consuming and subjective tuning process for tone mapping, tone mapping operators using deep learning are then proposed, which are often modeled as image-to-image translation tasks. 
Recently, tone-mapping methods using deep learning were also proposed. They are often modeled as image-to-image translation tasks. 
Yang~\etal~\cite{DRHT} use an autoencoder architecture to convert HDR images into LDR ones. However, artifacts like unreal color and contrast could be seen in their results. 
Rana~\etal~\cite{rana_deep_2019} use a multi-scale cGAN architecture. % and is trained with the same scale of image to generate LDR images. 
But it still suffers from halo artifacts and may result in other artifacts when test images are with a different scale from training images. 
Moreover, the above deep learning-based tone-mapping methods are all formulated as a one-to-one mapping and provide less variety of subjective styles.
%

%%%%%%%%%%%%%%%%%%%%%%%%%%%%%%%%%%%%%%%%%%%%%%%%%%%%%%%%%%%%%%%%%%%%%%%%%%%%%%%%%%%%
% THIS PAPER!!
%%%%%%%%%%%%%%%%%%%%%%%%%%%%%%%%%%%%%%%%%%%%%%%%%%%%%%%%%%%%%%%%%%%%%%%%%%%%%%%%%%%%
In this paper, a learning-based multimodal tone-mapping method is proposed.
%In this paper, a multi-mode, steerable, deep learning-based tone-mapping method is proposed.
%The proposed method can achieve visually appealing quality with minimal artifacts, outperforms most state-of-the-art tone mapping algorithms in both objective and subjective aspects, and runs in moderate processing speed. 
The proposed method could be divided into two parts. 
One is the \emph{EdgePreservingNet} that outputs locally varying kernels for decomposing the input HDR image into a base layer and a detail layer.
The other is \emph{ToneCompressingNet} that predicts the global tone compression curve.
Both of them run adaptively and dynamically according to the content and dynamic range of the input HDR image.

%\red{The multi-scale training data is incorporated to facilitate the robustness of our model. }
%A multi-scale training scheme is incorporated to facilitate the robustness of our model.
The proposed method achieves appealing quality with minimal artifacts, performs favorably against state-of-the-art tone-mapping algorithms both objectively and subjectively.
Furthermore, as a result of BicycleGAN~\cite{NIPS2017_6650} architecture, our method could generate diversified visually appealing tone-mapped results from a single HDR image. 

Our main contributions are summarized as follows:
\begin{enumerate}
    \item A deep learning-based tone-mapping method is proposed, which consists of an \emph{EdgePreservingNet} and a \emph{ToneCompressingNet}. 
    \item By integrating BicycleGAN architecture, the proposed method is able to generate various tone-mapped results from a single HDR image.
    \item By leveraging bilateral filters, the proposed method compresses the most part of dynamic range while preserves the high-frequency information of HDR images.
    %\item By integrating random latent codes during training, the proposed method is steerable in the testing phase.
    \item The proposed method performs favorably against existing methods in terms of both subjective and objective evaluations.

\end{enumerate}

\section{Related work}
\label{sec:related}
% \textbf{2.1 Tone mapping}\\
% \hspace*{1em} 
\subsection{Tone mapping}
% \heading{Conventional methods.}
%
There were many tone-mapping algorithms proposed in the past two decades. They could be roughly categorized into global methods and local methods depending on how the algorithm works. 
Global tone-mapping methods apply a single tone-mapping curve on each pixel in HDR images~\cite{Tumblin:1993:TRR:616030.617873,Ward:1994:CSL:180895.180934}, which often cause the loss of contrast and detail information.
In contrast, local tone-mapping methods utilize the spatial property to perform this task adaptively~\cite{bo_gu_local_2013}.
The global methods need less time to compute, while local methods generate better detail.
The local methods, in general, decompose images into two parts: a smooth base layer and a detail layer~\cite{Eilertsen:2017:CRT:3128975.3129025}.
Halo artifacts commonly occur around edges in local methods. Local tone-mapping algorithms are mainly proposed to reduce these artifacts.
Durand and Dorsey~\cite{durand_fast_2002} proposed using an edge-preserving bilateral filter for tone-mapping but halo artifacts still remained in some images.
Mantiuk et al.~\cite{Mantiuk:2006:PFC:1166087.1166095} proposed a contrast processing framework, but the detail was over-enhanced.
Farbman et al.~\cite{edg} proposed a multi-scale scheme using a weighted least square filter.
Liang et al.~\cite{liang_hybrid_2018} proposed a hybrid l1-l0 decomposition model.
%
%\newline \\
% \heading{CNN-based methods.} 

Although the previous works produce good results, hyper-parameter tuning is usually required to achieve the best visual quality and reduce halo artifacts for different images. 
Recently, deep learning-based methods were proposed without the requirement of parameter tuning and drastically reduced the computation time by utilizing powerful GPUs.
Patel et al. \cite{aga} use generative adversarial networks (GAN)~\cite{Goodfellow:2014:GAN:2969033.2969125} to perform tone-mapping.
However, the problem is over-simplified, and it could only be tested on 
%256 x 256 
$256 \times 256$ small patches.
Yang~\etal~\cite{DRHT} apply an autoencoder network with skip connections to transfer HDR images to LDR space. However, they fail to produce good results on general HDR images.
Rana et al.~\cite{rana_deep_2019} use conditional generative adversarial networks (cGAN)~\cite{DBLP:journals/corr/MirzaO14} and a multi-scale scheme to tone-map images.
Although the results achieve high TMQI~\cite{TMQI} scores, the results contain halo artifacts.
In this work, we adopt BicycleGAN~\cite{NIPS2017_6650} to allow our model to generate multiple high-quality tone-mapped images. The decomposition scheme enables our model to generate appealing results without halo effects.
%In this work, we use the BicycleGAN~\cite{NIPS2017_6650} to allow our model to generate multiple high-quality output images. The image decomposition scheme enables our model to generate good results with different sizes of inputs and without halo effect.
%

\subsection{Multimodal image-to-image translation}

Mode collapse is a well-known issue of cGAN~\cite{DBLP:journals/corr/MirzaO14}.
Bao et al.~\cite{DBLP:journals/corr/BaoCWLH17} proposed the cVAE-GAN, which combines a variational auto-encoder with a generative adversarial network that generates realistic and diverse results.
Zhu et al.~\cite{NIPS2017_6650} combine cVAE-GAN and cLR-GAN~\cite{DBLP:journals/corr/ChenDHSSA16,DBLP:journals/corr/DonahueKD16,Dumoulin2016AdversariallyLI} into BicycleGAN, which encourages the latent code generated by the encoder to be invertible and shows a better performance.
Yang et al.~\cite{DBLP:journals/corr/abs-1901-09024} proposed a novel regularization term in the generator to solve this mode collapse problem.  

\section{Method}
\label{sec:algorithm}
% \subsection{Bilateral Filtering for Tone Mapping} \label{sec:conv-pipeline}
\subsection{Learning-based Bilateral Filters} \label{our-pipeline}
%
% \begin{figure*}[]
%     \centering
%     \includegraphics[width=\textwidth]{images/conventional_pipeline/conventional_pipeline.pdf}
%     \caption{\textbf{Bilateral Filtering for Tone Mapping.} Given an input HDR image, the luminance is extracted first and passed to an edge-preserving filter to generate the base layer. Then, the detail layer is acquired by computing the difference. The base layer is compressed using a fixed global tone-curve, such as exponential or sigmoid function. On the contrary, the detail layer is enhanced and added back with the base layer after the mapping. This step not only preserves the high-frequency details but also compresses the dynamic range of the main part of the HDR image. Finally, some post-processing and color restoration is performed in order for display}
%     \label{fig:conventional_pipeline}
% \end{figure*}
%
%
\begin{figure*}[]
    \centering
    \includegraphics[width=\textwidth]{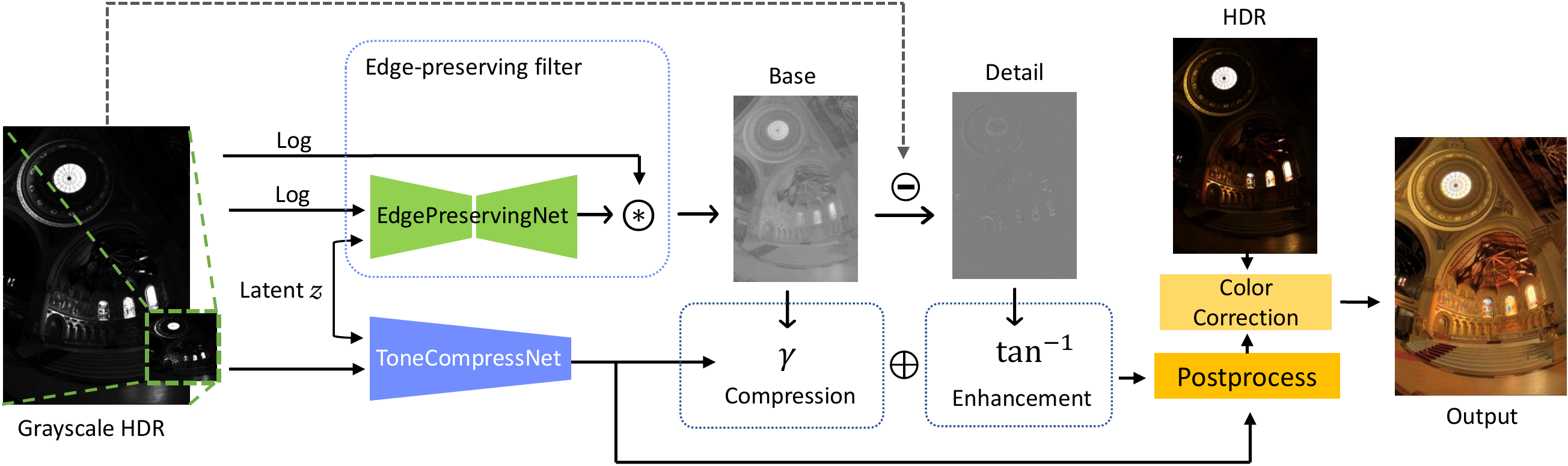}
    \caption{\textbf{An overview of our tone-mapping method.} Our proposed method comprises two networks: (a) an \emph{EdgePreservingNet}, and (b) a \emph{ToneCompressingNet}. Given an input HDR image in logarithm domain, the \emph{EdgePreservingNet} generates convolutional kernels, which will be used to produce the base image. Next, we subtract the base image from the input HDR image to acquire the detail image, and then apply an enhancement operation to it. The base image is then compressed using the global tone curve predicted by the \emph{ToneCompressingNet}. Eventually, the output LDR image is obtained by adding the compressed base image and enhanced detail image along with some post-processing and color correction operations.}
    \label{fig:generator}
\end{figure*}

Bilateral filtering \cite{durand_fast_2002} is one of the most common tone-mapping operators. The core concept behind this operator is to decompose an HDR image into a base layer and a detail layer, which respectively stands for the most part of dynamic range and the high-frequency information of the HDR image.
%Thus, a detail-preserved LDR image with high compression ratio could be obtained by applying compression operations only to the base layer.
However, the base layer and detail layer are typically decomposed by some hand-crafted edge-preserving filters and compression operations. Due to the large amount of parameters, tuning these filters and operations is usually difficult and time-consuming.

% \subsection{Learning-based Bilateral Filtering} \label{our-pipeline}

\begin{figure}
\centering
\renewcommand{\tabcolsep}{1pt} % adjust horizontal space
\renewcommand{\arraystretch}{1} % adjust vertical space
\begin{tabular}{cc}
    \includegraphics[width=.45\linewidth]{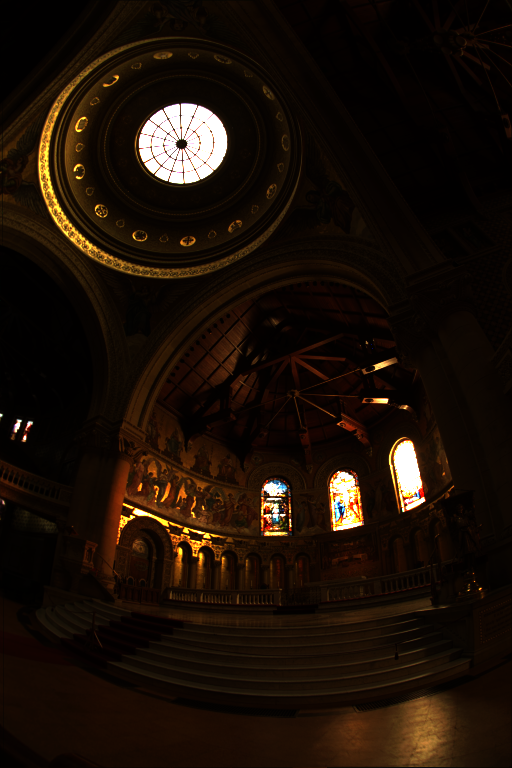} & 
    \includegraphics[width=.45 \linewidth]{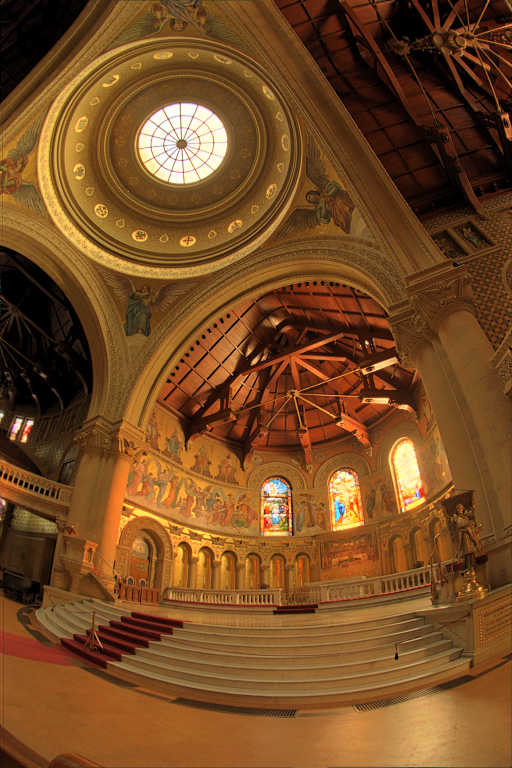} \\
    (a) Input HDR image & (b) Our LDR result  \\
\end{tabular}
\begin{tabular}{cc}
    \includegraphics[width=.45\linewidth]{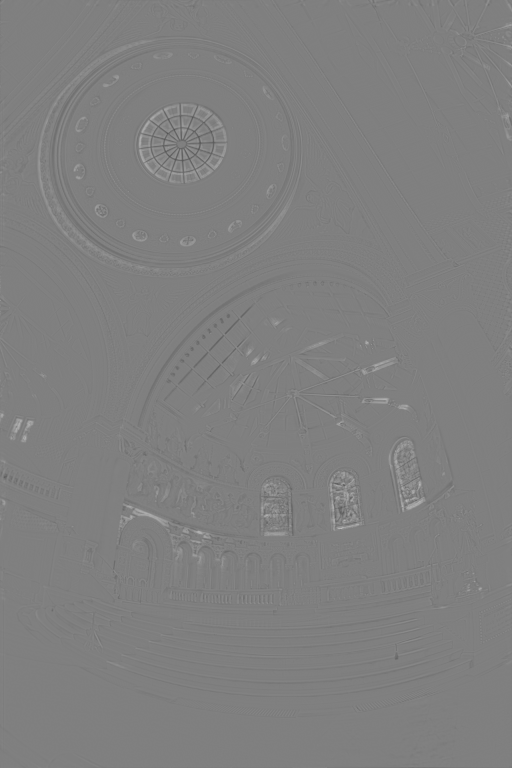} & 
    \includegraphics[width=.45 \linewidth]{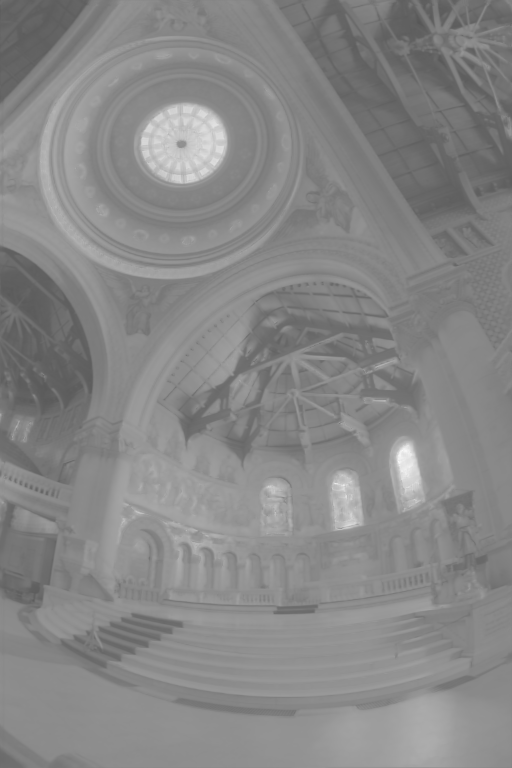} \\
    (c) Learned detail image & (d) Learned base image \\
\end{tabular}
\caption{\textbf{Example of learned decomposed images.}}
\label{fig:base_detail}
\end{figure}

Instead of hand-crafted filters and operations, we propose a learning-based scheme as shown in Fig.~\ref{fig:generator}. The proposed scheme comprises two networks: (a) an \emph{EdgePreservingNet}, and (b) a \emph{ToneCompressingNet}. In order to avoid artifacts, we let \emph{EdgePreservingNet} be a Kernel Prediction Network (KPN)~\cite{DBLP:journals/corr/abs-1712-02327}. Therefore, given an input HDR image in logarithm domain, the \emph{EdgePreservingNet} generates convolutional kernels to produce the base image. Next, the detail image is acquired by subtracting the base image from the input HDR image. For the purpose of better visual quality, we apply an enhancement operation to the detail image. The base image is then compressed using the global tone curve predicted by the \emph{ToneCompressingNet}, which is a typical Conv-FC network. Eventually, the output LDR image is obtained by adding the compressed base image and enhanced detail image along with some post-processing and color correction operations. Fig.~\ref{fig:base_detail} shows an example of decomposed images.

The \emph{EdgePreservingNet} and \emph{ToneCompressingNet} are jointly trained using the framework of BicycleGAN, which will be described in Section~\ref{bicycle-arch}. It is worthwhile to mention that various random latent codes $\mathbf{z}$ are fed into these networks during training to make them able to generate various LDR images. Moreover, a latent code optimization scheme will be introduced in Section~\ref{sec:latentcode_select} to help users to find appropriate latent codes.

% A latent code $z$ is fed into both EdgePreservingNet and ToneCompressingNet to control the tone-mapped results.
%
% It is worth mentioning that all the users need to do is \red{to determine the input latent code}.
%
% \red{The users can also adjust the input latent code to obtain different tone mapped images.}
%
% Moreover, a scheme utilizing the state-of-the-art tone-mapping objective quality index TMQI~\cite{TMQI} to automatically pick the latent code candidates will be introduced in Section~\ref{sec:latentcode_select}.
%A latent code $z$ is fed into both EdgePreservingNet and ToneCompressingNet to control the tone-mapped results. It is worth mentioning that all the users need to do is to determine this latent code, which makes parameter tuning much easier. Moreover, a scheme utilizing the state-of-the-art tone-mapping objective quality index TMQI~\cite{TMQI} to automatically pick the latent code candidates will be introduced in Section~\ref{sec:quantitative}.
% For network architecture, we choose Kernel Prediction Network (KPN)~\cite{DBLP:journals/corr/abs-1712-02327} for EdgePreservingNet and a typical Conv-FC architecture for ToneCompressingNet. These two networks are integrated into the framework of an improved cVAE-GAN, which will be described in Section~\ref{bicycle-arch}.
%

% Fig.~\ref{fig:base_detail} shows an example of our learned base/detail layer decomposition.
%

\subsection{Tone Mapping Operators}\label{generator}
\begin{figure}
% \begin{minipage}[t]{0.45\linewidth}
    \centering
    \renewcommand{\tabcolsep}{1pt} % adjust horizontal space
    \renewcommand{\arraystretch}{1} % adjust vertical space
    \begin{tabular}{cc}
        \includegraphics[width=.4\linewidth]{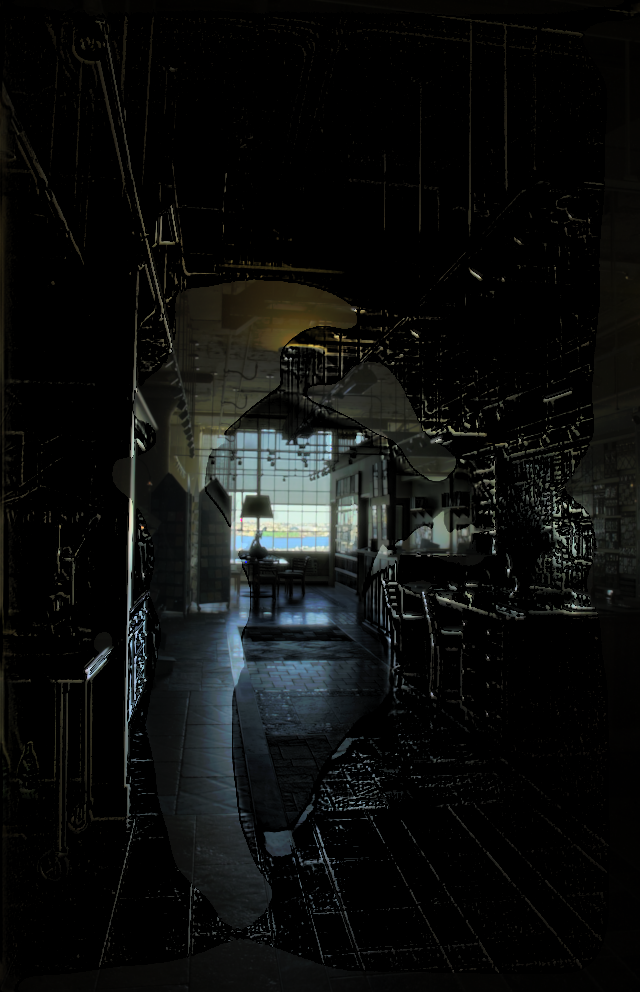} & 
        \includegraphics[width=.4 \linewidth]{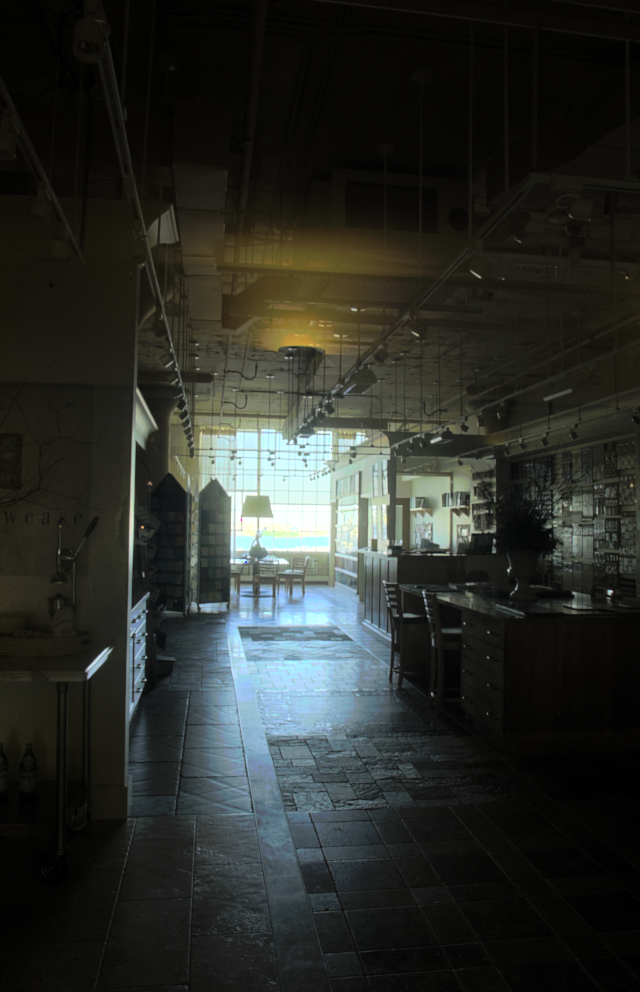} \\
        (a) without & (b) with  \\
    \end{tabular}
    \caption{\textbf{The effect of normalization on predicted kernels.}}
    \label{fig:test}
    \end{figure}

% \heading{Generative model pipeline.} 
% Let $X$ and $Y$ denote the HDR image domain and LDR image domain, respectively. Our goal is to learn the generator $G$, which \red{transforms the image from HDR domain $X$ to LDR image domain $Y$}, that is, $X\xrightarrow{G}Y$.
%Let $X$ and $Y$ denote the HDR image domain and LDR image domain, respectively. Our goal is to learn the generator $G$, which transforms the image of HDR domain image $X$ to LDR image domain $Y$, that is, $X\xrightarrow{G}Y$.
% This generator can be divided into two parts: \red{EdgePreservingNet and ToneCompressingNet} with different goals. \red{EdgePreservingNet} predicts the base image as the part of the edge-preserving filter in the conventional method, while the \red{ToneCompressingNet} is used to predict the base compression rate and the degree of post-processing.
%This generator can be divided into two parts: local path and global path with different goals. Local path predicts the base image as the part of the edge-preserving filter in the conventional method, while the global path is used to predict the base compression rate and the degree of post-processing.
We apply a U-Net~\cite{UNet} architecture to the \emph{EdgePreservingNet}, which is composed of an encoder-decoder with some skip connections. As suggested in Gu \etal~\cite{bo_gu_local_2013}, the input HDR images are first transformed into logarithm domain and then normalized to $[0, 1]$ to fit human perception. Rather than directly generating the base image, the \emph{EdgePreservingNet} predicts a pixel-wise filter of size $H \times W \times K^2$, where $K$ is the kernel size, and $H$, $W$ is the height and width of the image. The predicted kernel at each pixel $\mathbf{w}_{p}$ is then normalized by
\begin{equation}
\mathbf{w}_{p}' = \frac{\mathbf{w}_{p}}{\sum_{i=1}^{K^2} \mathbf{w}_{p}(i)},
\label{eq:1}
\end{equation}
where $i$ is termed as each element of $\mathbf{w}_{p}$. Fig.~\ref{fig:test} shows that this normalization is critical to the tone-mapping performance.

The base image $\mathbf{I}_\text{base}$ is thus given by applying convolution to the input HDR image $\mathbf{I}_\text{hdr}$ in logarithm domain as

\begin{equation}
{\mathbf{I}_\text{base}} = \sum_{p=1}^{H \times W}(\mathbf{w}_{p}' * \log{\mathbf{I}_\text{hdr}}) \cdot \mathds{1}_p,
\label{eq:B}
\end{equation}
where $\mathds{1}_p$ indicates pixel $p$ having the value $1$ otherwise $0$. The detail image is then obtained by $\mathbf{I}_\text{detail} = \mathbf{I}_\text{hdr} - \mathbf{I}_\text{base}$.

%\begin{equation}
%\mathbf{I}_\text{detail} = \mathbf{I}_\text{hdr} - %\mathbf{I}_\text{base}.
%\label{eq:D}
%\end{equation}

%The resulting base layer image $B_{p}$ is reconstructed as a linear combination of pre-processed grayscale input $I$ and the normalized edge-preserving kernels $w_{p}'$ in a $K \times K$ neighborhood $N(p)$ around image pixel $p$.
%The detail layer $D_{p}$ is then obtained by subtracting the base layer from the input $I_{p}$.

%\begin{gather}
%B_p = \sum_{q \in N(p)} w_p'(q) * I_q \label{eq:B}
%\\
%D_p = I_p - B_p  \label{eq:D}
%\end{gather}
%

The \emph{ToneCompressingNet} consists of a series of consecutive convolution layers followed by some fully-connected layers. The network predicts the compression rate $\gamma_\text{base}$ as well as the degree of post-processing $\gamma_\text{post}$. The compressed base image is simply acquired by
\begin{equation}
{\mathbf{I}_\text{base}}' = {\mathbf{I}_\text{base}}^{\gamma_\text{base}}.
\end{equation}

%An HDR image with a wider dynamic range should be compressed more. So the linear HDR image is used as input, which preserves the dynamic range of HDR image; the model could learn the corresponding compression gamma rate according to the HDR image. In order to handle different sizes of input, the HDR images are resized to $256 \times 256$ \red{and normalized with the min/max values in the original full-resolution HDR image.} The resizing process does not change the dynamic range of the image.

%

%
To better preserve visual information, we define our enhancement function as
\begin{equation}
E(x|\alpha ) = \frac{\arctan \alpha x}{\arctan \alpha} \label{eq:expansion},
\end{equation}
where $\alpha$ is a hyper-parameter controlling the enhancement level. Specifically, $\alpha$ is set to $3.5$ in this work, so we have the enhanced detail image ${\mathbf{I}_\text{detail}}' = E(\mathbf{I}_\text{detail}|\alpha = 3.5)$.
The reconstructed image is thus given by $\mathbf{I}_\text{rec} = {\mathbf{I}_\text{base}}' + {\mathbf{I}_\text{detail}}'$.
%\begin{equation}
%\mathbf{I}_\text{rec} = {\mathbf{I}_\text{base}}' + {\mathbf{I}_\text{detail}}'.
%\label{equ:I'}
%\end{equation}

To further enhance the contrast of $\mathbf{I}_\text{rec}$, we stretch and flatten its histogram by Eq.~\ref{eq:expansion} again.
However, different from detail enhancement, this operation needs to be applied on zero-mean domain. Thus, the post-processing operation is defined as
\begin{equation}
\mathbf{I}_\text{post} = E(\mathbf{I}_\text{rec} - \mu|\alpha = \gamma_\text{post}) + \mu,
\label{equ:postprocess}
\end{equation}
where $\mu$ is the mean illuminance of $\mathbf{I}_\text{rec}$.

Finally, following Tumblin \etal~\cite{Tumblin:1999:LBH:311535.311544}, we perform a color correction to obtain our LDR image as
\begin{equation}
\mathbf{I}_\text{ldr} = (\frac{\mathbf{I}_\text{hdr}^{c}}{\mathbf{I}_\text{hdr}})^{\beta} \cdot \mathbf{I}_\text{post} \label{eq:color-correction},
\end{equation}
where $\mathbf{I}_\text{hdr}^{c}$ denotes the color HDR image (original radiance map) and $\beta$ is set to 0.6 in this work.

\subsection{Training}\label{bicycle-arch} 
% BicycleGAN architecture 
% contains three parts: the generator discussed above, the discriminator, and the encoder. We use the PatchGAN~\cite{DBLP:journals/corr/IsolaZZE16} as the discriminator, which splits the image into many patches and discriminates each patch is real or fake. The encoder is used to generate the latent code 
% $z$ and help the network produce different results.
% \newline \\
%\heading{Objective function.}

We choose BicycleGAN as our main framework to train tone-mapping operators.
Nevertheless, we replace the LSGAN loss used in BicycleGAN with a hinge loss, which has been proven to produce impressive results in~\cite{DBLP:journals/corr/abs-1809-11096}.
The hinge loss can be expressed as
\begin{equation}
\begin{split}
\mathcal{L}_{G} &= -\mathbb{E}_{\mathbf{x},\mathbf{y}\sim \mathbb{P}_\text{data},\mathbf{z}\sim \mathbb{P}_\mathbf{z}}[D(G(\mathbf{x},\mathbf{z}),\mathbf{y})] \\
\mathcal{L}_{D} &= -\mathbb{E}_{\mathbf{x},\mathbf{y}\sim \mathbb{P}_\text{data}}[min(0,-1 + D(\mathbf{x},\mathbf{y}))] \\
&-\mathbb{E}_{\mathbf{x},\mathbf{y}\sim \mathbb{P}_\text{data},\mathbf{z}\sim \mathbb{P}_\mathbf{z}}[min(0,-1 - D(G(\mathbf{x},\mathbf{z}), \mathbf{y}))],
\end{split}
\end{equation}
where $\mathcal{L}_{G}$ is the generator loss and $\mathcal{L}_{D}$ is the discriminator loss. The $\mathbf{x}$ and $\mathbf{y}$ here represent the LDR images in target domain. 

To further improve the diversity, a regularization term $\mathcal{L}_\text{div}$ proposed by DSGAN~\cite{DBLP:journals/corr/abs-1901-09024} is introduced as
\begin{equation}
\mathcal{L}_\text{div} = \mathbb{E}_{\mathbf{z}_1,\mathbf{z}_2\sim\mathbb{P}_\mathbf{z}} \left [ min (\frac{\left \| G(\mathbf{x},\mathbf{z}_1)-G(\mathbf{x},\mathbf{z}_2) \right \|_1}{\left \| \mathbf{z}_1-\mathbf{z}_2 \right \|_1}, \tau ) \right ],\\
\end{equation}
where $\tau$ determines the stability of numerical computation. Our GAN loss is hence defined as
\begin{equation}
\mathcal{L}_\text{GAN}(G,D) = \mathcal{L}_{G} + \mathcal{L}_{D} - \lambda_\text{div} \mathcal{L}_\text{div}.
\end{equation}

In addition, a total variation loss $\mathcal{L}_\text{tv}$ is imposed on the base image to smooth the prediction as
\begin{equation}
\mathcal{L}_\text{tv} = \mathbb{E}_{\mathbf{I}_\text{base} \sim \mathbb{P}_g} \left [ \sum_{p=1}^{H \times W} \left | \nabla {\mathbf{I}_\text{base} (p)} \right | \right ],
\end{equation}
where $\left | \nabla \cdot \right |$ stands for the magnitude of image gradient.

The full objective of our model is thus combined as
\begin{equation}
\mathcal{L}_\text{total} =
\mathcal{L}_\text{GAN}(G,D) + \lambda_\text{rec} \mathcal{L}_\text{rec}
+\mathcal{L}_\text{kl}+\mathcal{L}_z+\mathcal{L}_\text{tv},
\end{equation}
where $\mathcal{L}_{rec}$, $\mathcal{L}_{kl}$, and $\mathcal{L}_z$ respectively represent the $L_1$ reconstruction loss, the KL divergence of latent codes, and the encoder loss used in the original BicycleGAN.
%the $\mathcal{L}_1$ reconstruction loss of KPN filter-based generator $\mathcal{L}_{rec}$ 

%%%

% \end{minipage}%
    % \hfill%

\begin{figure}
\centering
\footnotesize
\begin{tabular}{c}
    \includegraphics[width=0.9\linewidth]{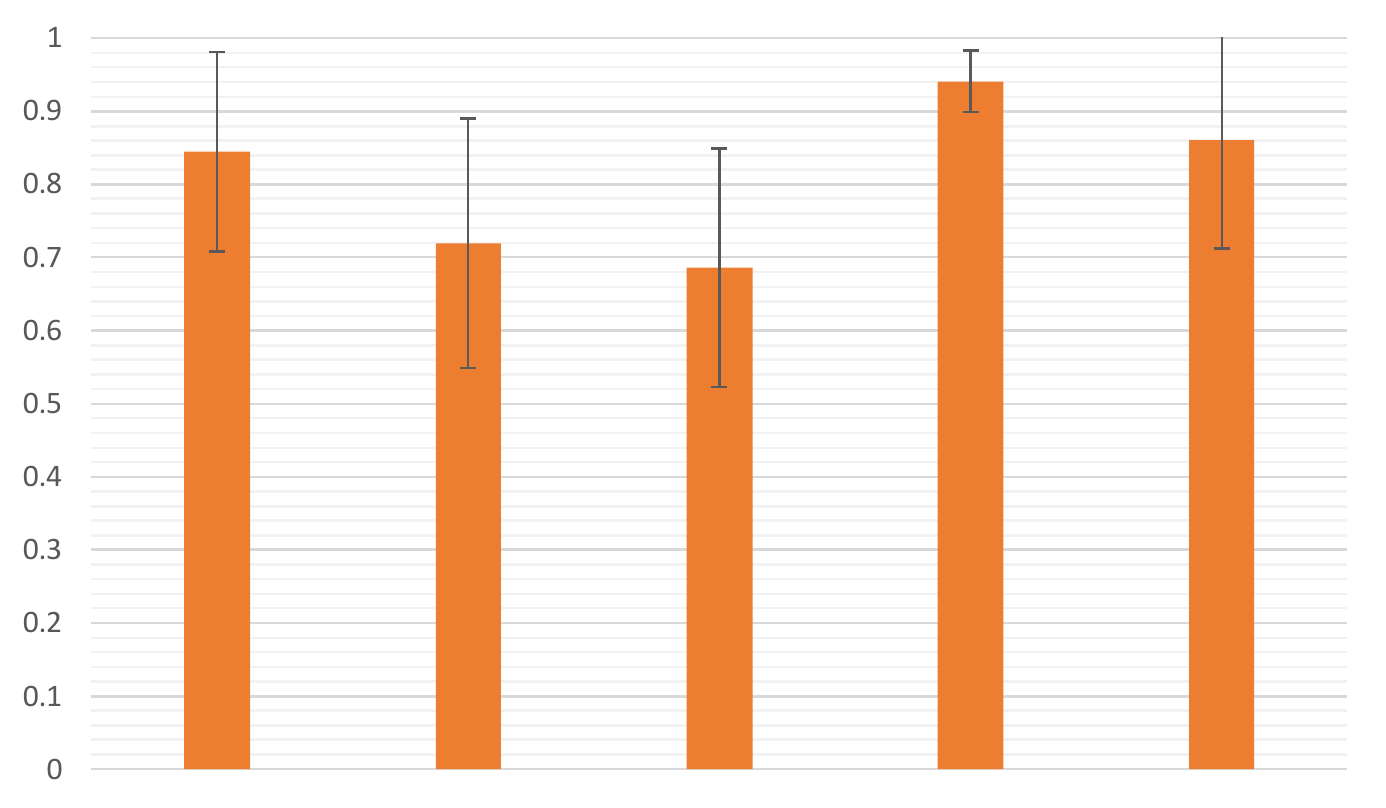} \\
\end{tabular}
\resizebox{0.9\linewidth}{!}{
\begin{tabular}{cccccc}
\enspace \enspace \enspace \enspace \enspace  & \makecell{vs.\\ Liang~\etal~\cite{liang_hybrid_2018}} & \makecell{vs.\\ Mantiuk~\etal~\cite{Mantiuk2008}} & \makecell{vs.\\ Lischinski~\etal~\cite{Lischinski}} & \makecell{vs.\\ Mantiuk~\etal~\cite{Mantiuk:2006:PFC:1166087.1166095}} & \makecell{vs.\\ Yang~\etal~\cite{DRHT}} \\
\end{tabular}
}
\caption{\textbf{User study results.} Our results achieve higher probability of being selected by the subjects than the compared methods.}
\label{fig:user_study}
\end{figure}

\subsection{Latent Code Optimization} \label{sec:latentcode_select}
Recall that tone-mapping is a subjective task, that is, people prefer different types or styles of tone-mapped images. Although our method allows users to change the style by adjusting the latent code, it is still very challenging to find appropriate ones because of the extremely huge search space. Instead of using random latent codes in testing phase, we propose a scheme that optimizes TMQI~\cite{TMQI}, which is a representative evaluation metric of tone-mapping, to help users to filter out inappropriate latent codes. Given a well-trained tone-mapping operator with fixed model parameters and an initial latent code, the latent code is then iteratively optimized by backpropagation using Adam~\cite{adam} optimizer. Generally, this process converges by about 30 iterations. With this scheme, all the users need to do is select latent codes among few candidates. Note that both TMQI and our model are differentiable.

\section{Experimental Results}
\label{sec:experiments}
% \su{the testing dataset is come from one dataset :Fairchild dataset(105 images in total) } ok
% wow good good
% \textcolor{magenta}{
The following experiments are conducted on Fairchild's dataset~\cite{Fairchild}.
This test set contains 105 images with diverse scenes including indoor, outdoor, daylight, and night views with a variety of image sizes.
%
%The steerable ability of our model is demonstrated by changing the latent codes. 
%
%\red{The proposed method is able to generate results in various styles with different latent codes as shown in Fig.~\ref{fig:teaser}, and the strategies mentioned in Sec.\ref{sec:latentcode_select} help users to find a decent one among them.}
%The proposed method can generate results in various styles with different latent codes, and the property of the results changes progressively as the latent code walks through the interpolated space. 
%The general performance of our model is evaluated with zero latent codes. 
%
We compare our method with state-of-the-art tone-mapping methods qualitatively and quantitatively.
Additionally, we provide a user study that also shows the superiority of our method over the compared methods.

\subsection{Training Dataset} \label{subsec:training_data} % this will be refered in experiment

We collect our training dataset from multiple sources~\cite{HDRheaven,Fairchild,HDReyes,Nemoto:203873,HDRdb,SYNS,HDRlab,Debevec97recoveringhigh,HDRSID}, which totally contains 1032 HDR images with a wide range of contents, e.g. scenes, cameras, and shooting settings.
%Different TMO algorithms are suitable in their specialized type of images, so 16 TMO algorithms are selected with the default parameters in their original literature 
%
%\cite{liang_hybrid_2018,Mantiuk:2006:PFC:1166087.1166095,Mantiuk2008,fattal_gradient_2002,Ferradans,drago,durand_fast_2002,Reinhard05,reinhard_photographic_2002,Ashikhmin,Pattanaik,mai,Ferradans,Kim_Kautz:2008a,VanHateren,Lischinski,Ferwerda:1996:MVA:237170.237262}
%
We use Luminance HDR~\footnote{http://qtpfsgui.sourceforge.net/} with the default parameters to generate LDR images, and the LDR images corresponding to each HDR image with top-3 highest TMQI scores are selected as the target images. This selection scheme makes our model has better generalization. Note that our training dataset and Fairchild's dataset are mutually exclusive.

\subsection{Implementation details}

We apply random cropping and flipping augmentation to our training data. During training, the images are cropped into $256 \times 256$ patches. The initial learning rates are set to $5 \cdot 10^{-5}$ and $ 2 \cdot 10^{-4}$ for the tone-mapping operator and the discriminator, respectively. The model parameters are initialized by Xavier initializer~\cite{pmlr-v9-glorot10a} and optimized by Adam optimizer with $\beta_{1}=0.9$ and $\beta_{2}=0.998$. The kernel size $K$ for \emph{EdgePreservingNet} is set to $7$. The valid range of $\gamma_\text{base}$ and $\gamma_\text{post}$ are set to $[0.8, 2.8]$ and $[1.7, 3.7]$, respectively.
The hyper-parameters $\lambda_\text{rec}$ and $\lambda_\text{div}$ are respectively set to 1 and 5. The training process takes about 1 day for 300 epochs on a single NVIDIA TITAN RTX GPU.

\subsection{Qualitative Comparison}
\label{sec:qualitative}
\begin{figure*}[htb]
\centering
\renewcommand{\tabcolsep}{1pt} % adjust horizontal space
\renewcommand{\arraystretch}{1} % adjust vertical space

\begin{tabular}{cccc}
    \includegraphics[width=.25\linewidth]{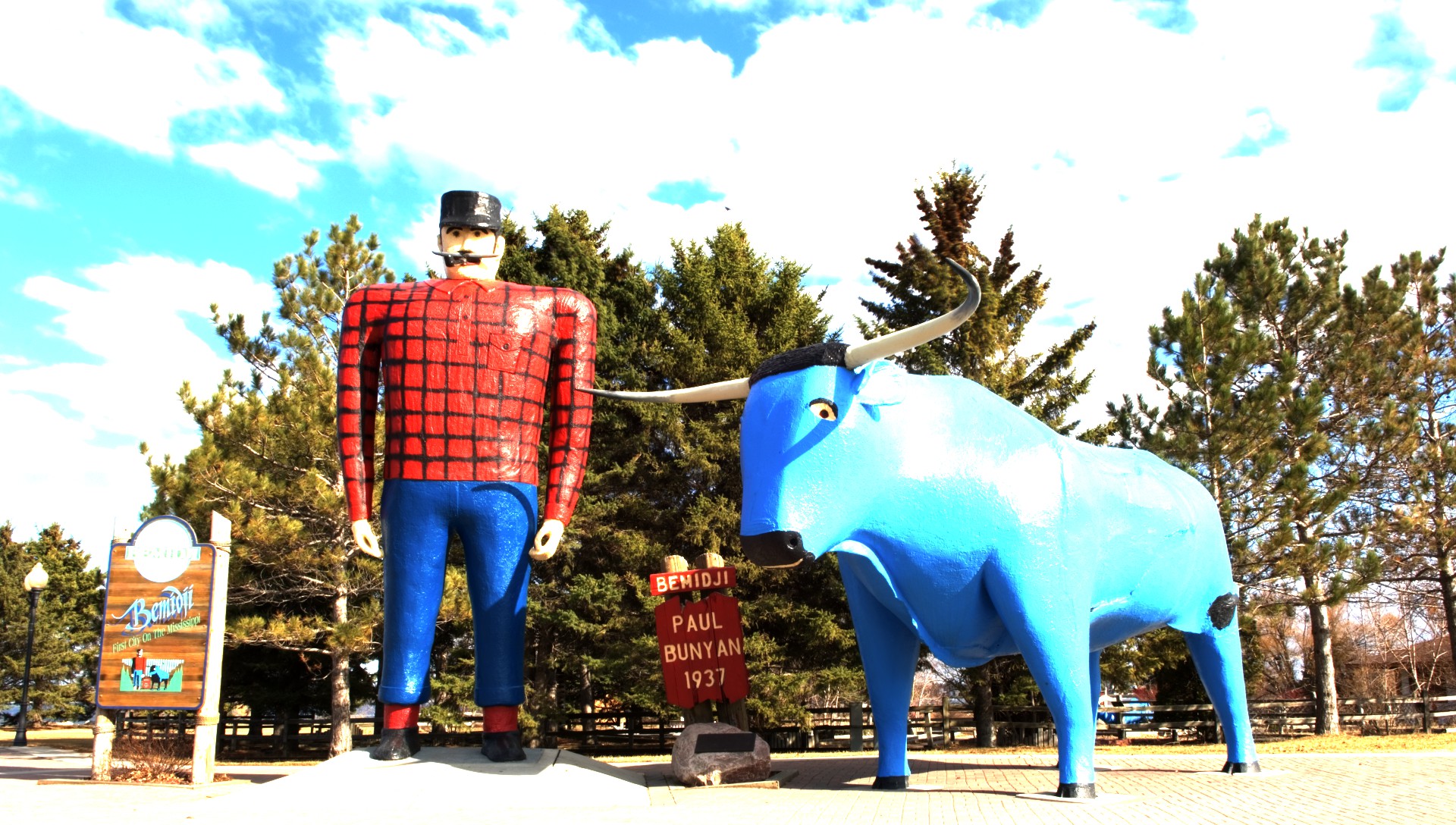} & \includegraphics[width=.25\linewidth]{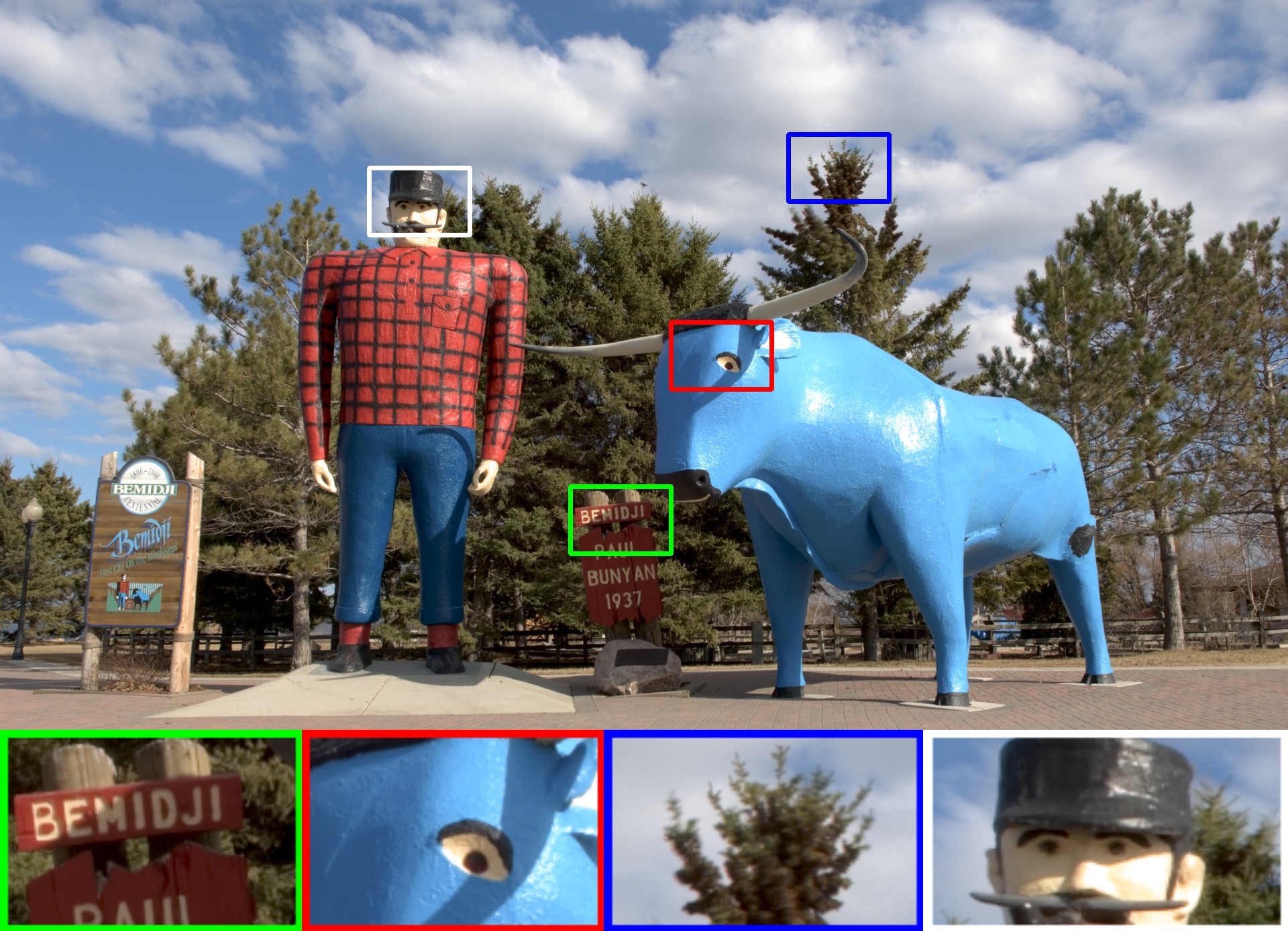} & \includegraphics[width=.25\linewidth]{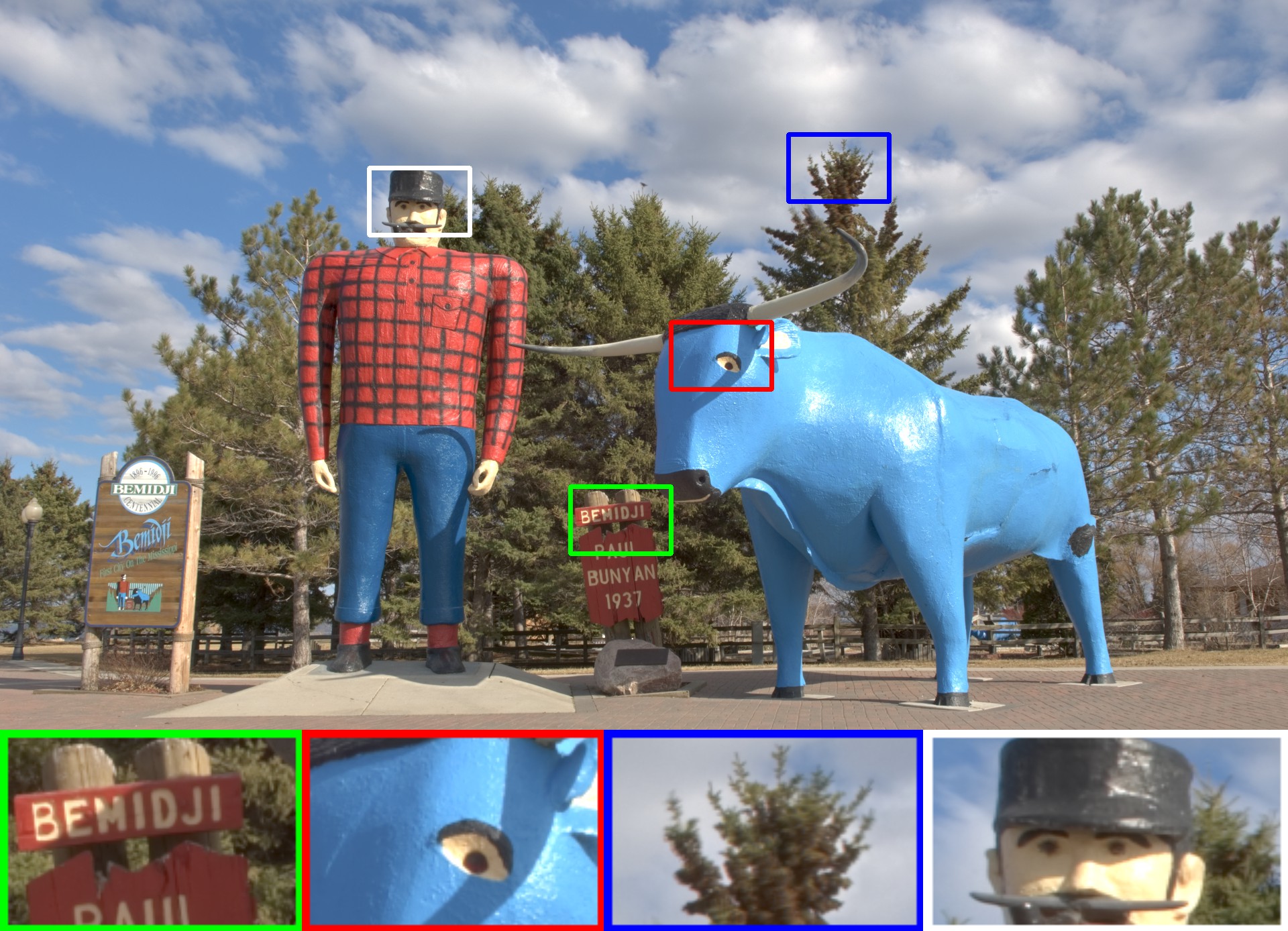} & \includegraphics[width=.25\linewidth]{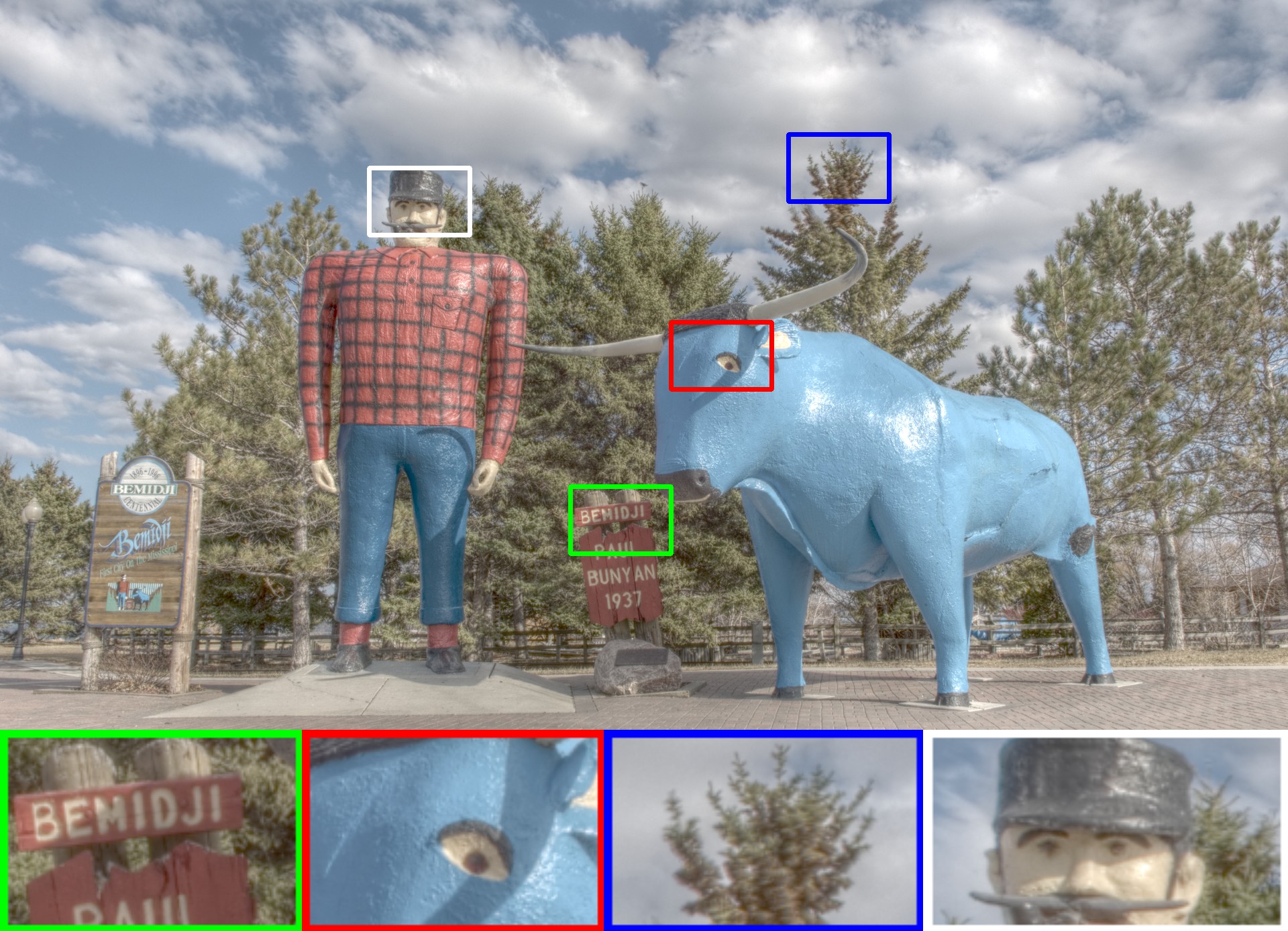} \\
    (a) linear & (b) Mantiuk~\etal~\cite{Mantiuk2008} & (c) Lischinski~\etal~\cite{Lischinski} & (d) Mantiu~\etal~\cite{Mantiuk:2006:PFC:1166087.1166095} \\
\end{tabular}

\begin{tabular}{cccc}
    \includegraphics[width=.25\linewidth]{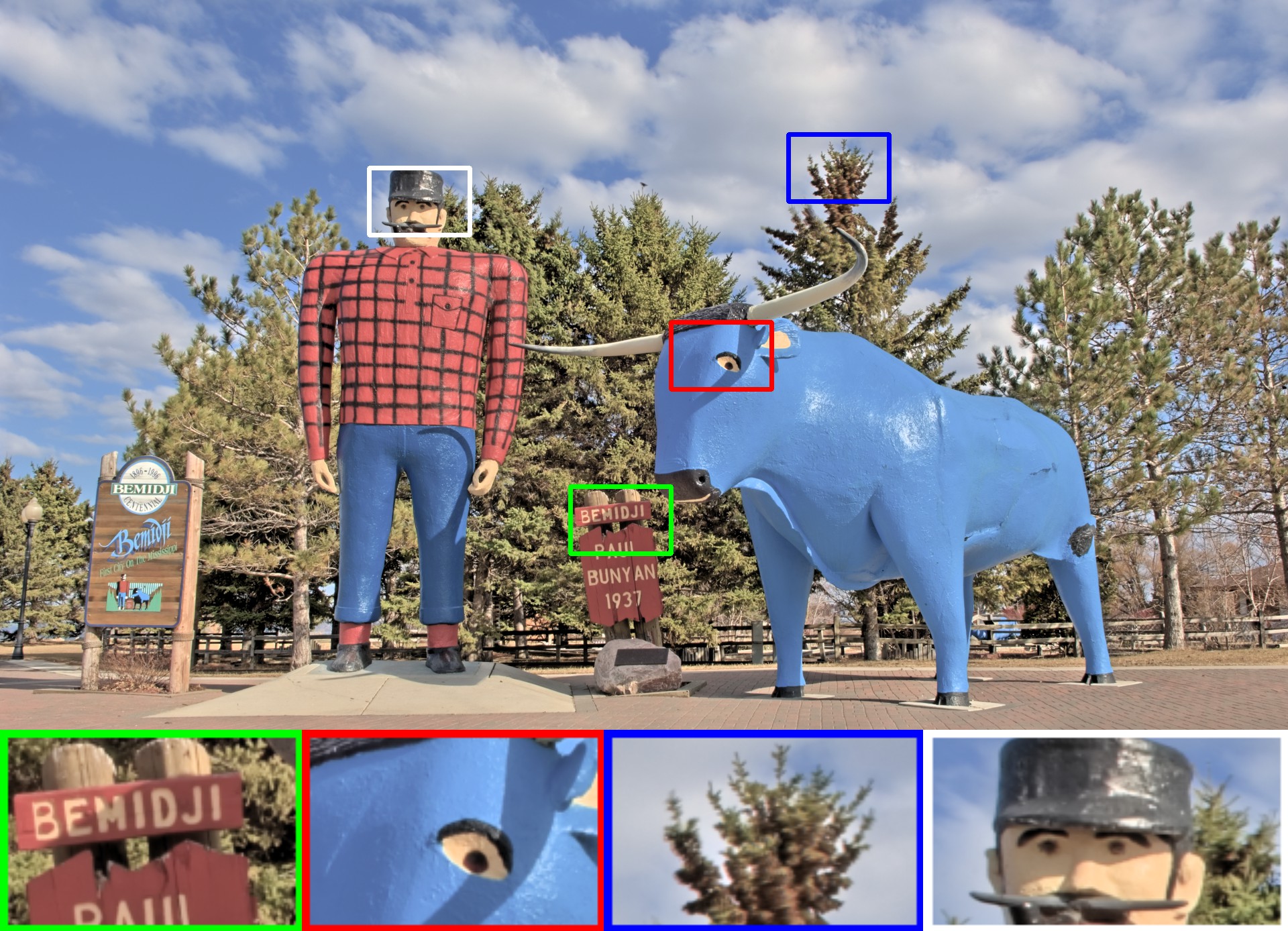} & \includegraphics[width=.25\linewidth]{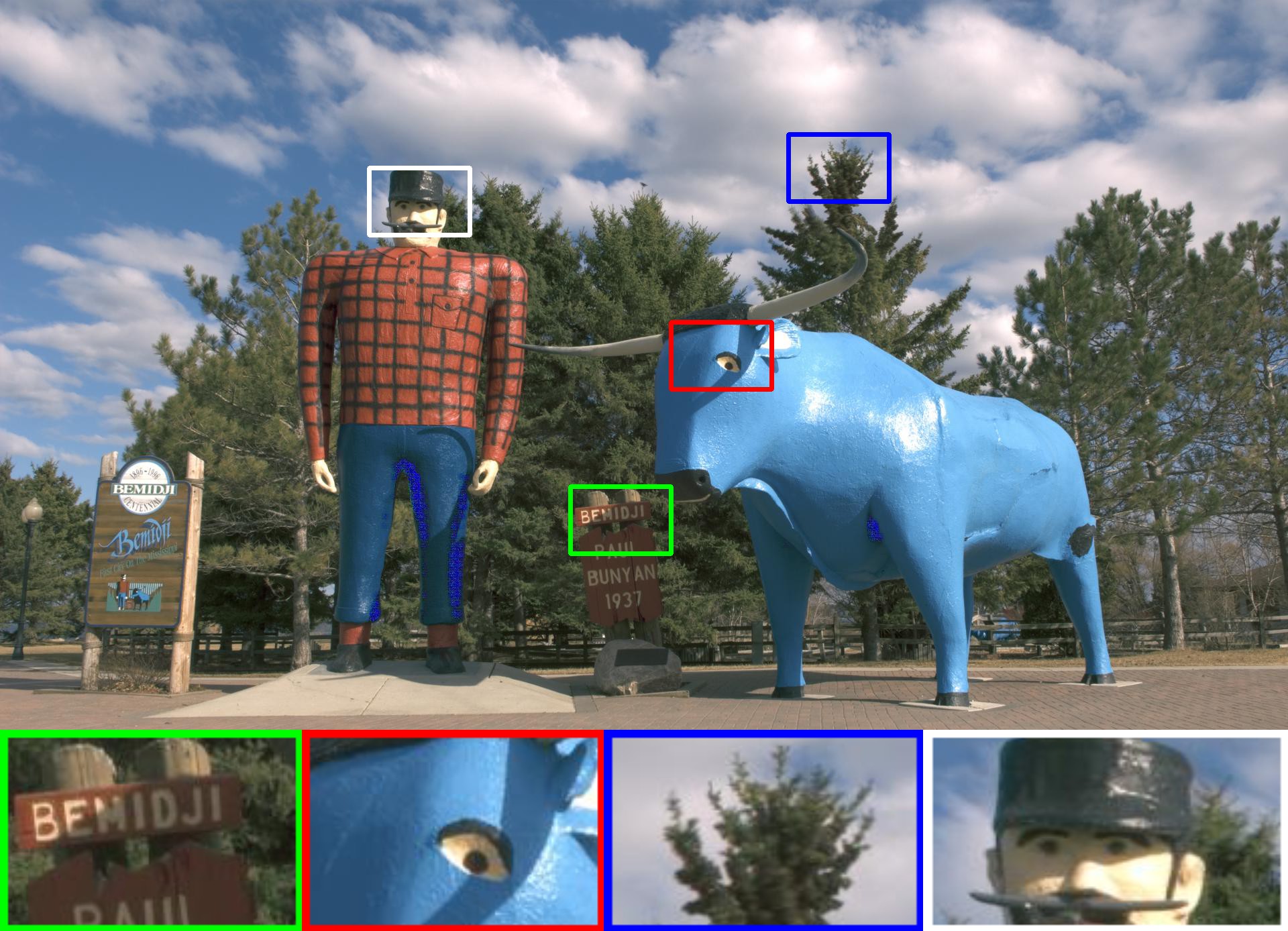} & \includegraphics[width=.25\linewidth]{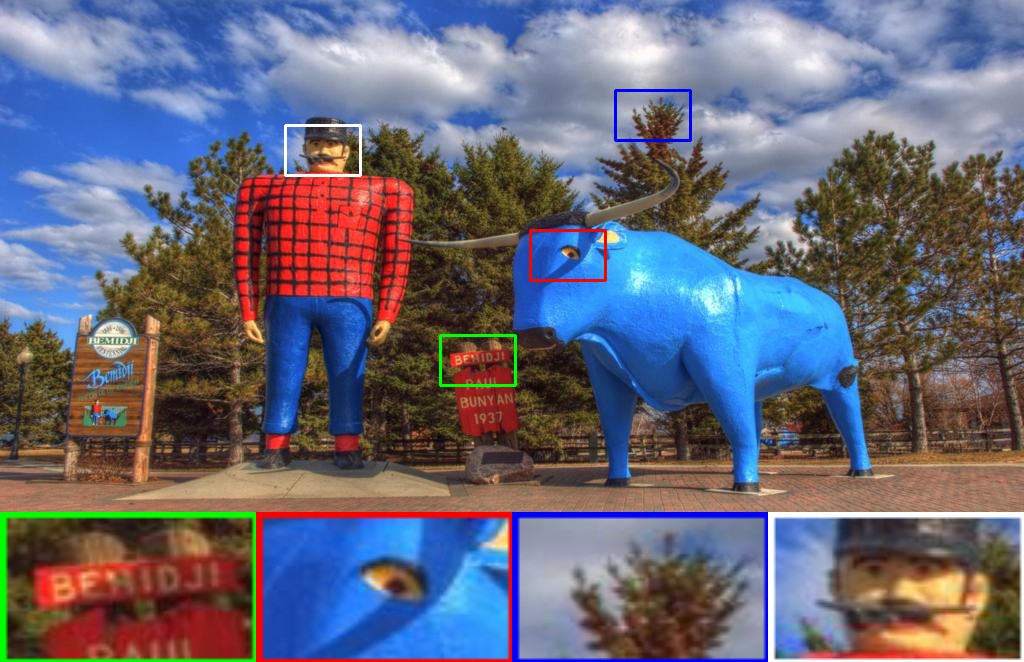} & \includegraphics[width=.25\linewidth]{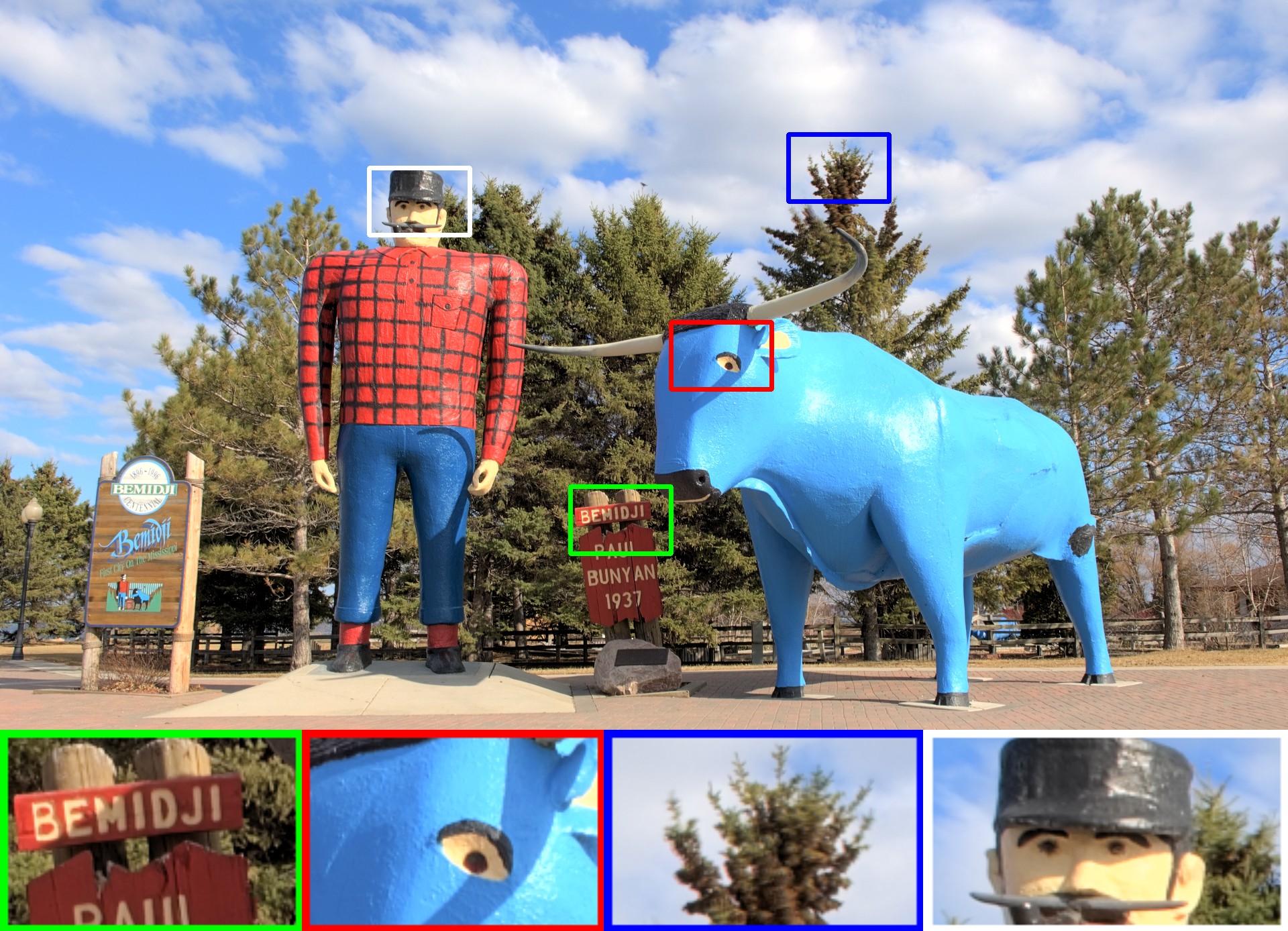} \\
    (e) Liang~\etal~\cite{liang_hybrid_2018}& (f) Yang~\etal~\cite{DRHT} & (g) *Rana~\etal~\cite{rana_deep_2019} & (h) Ours \\
\end{tabular}

\begin{tabular}{cccc}
    \includegraphics[width=.25\linewidth]{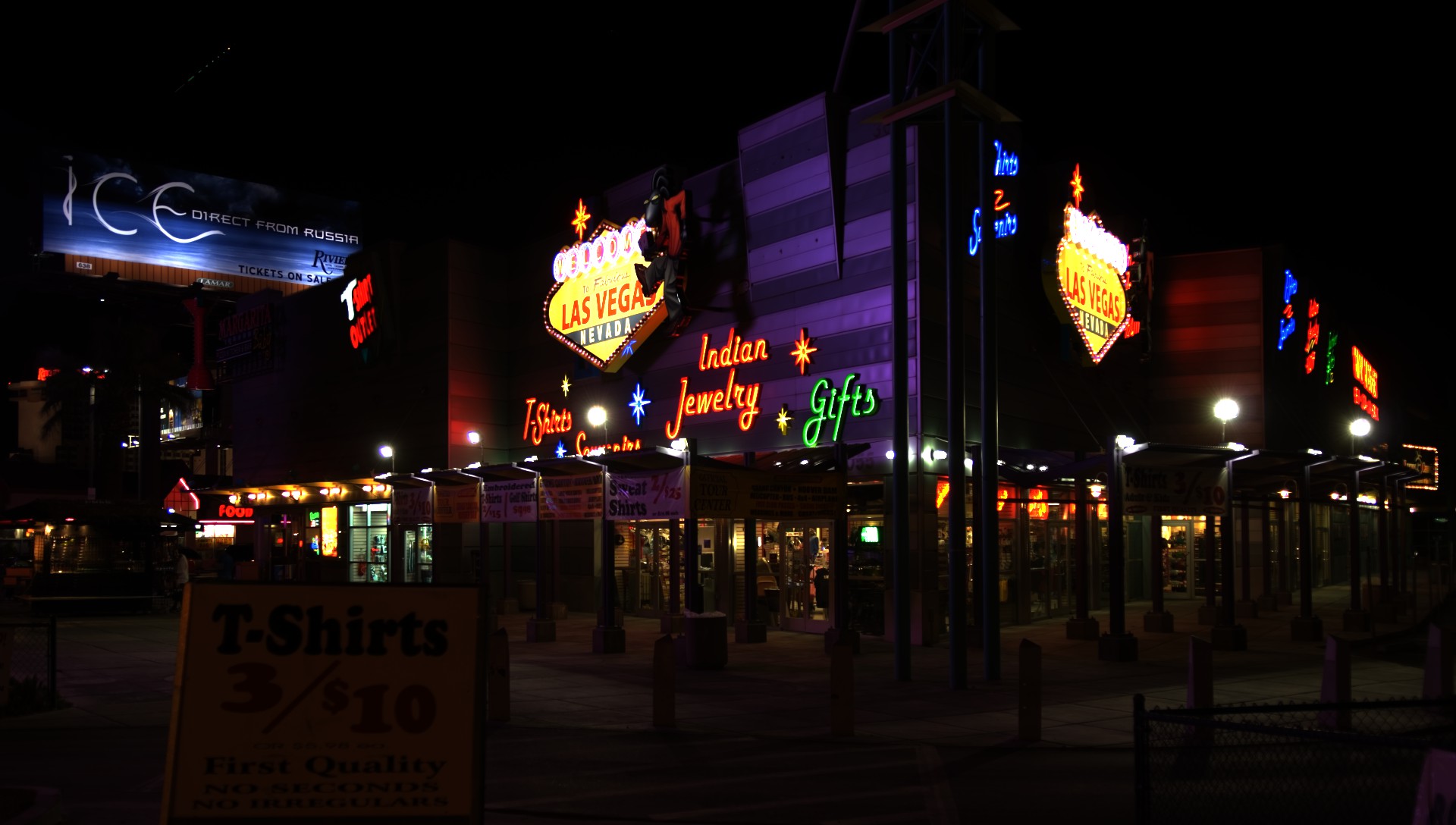} & \includegraphics[width=.25\linewidth]{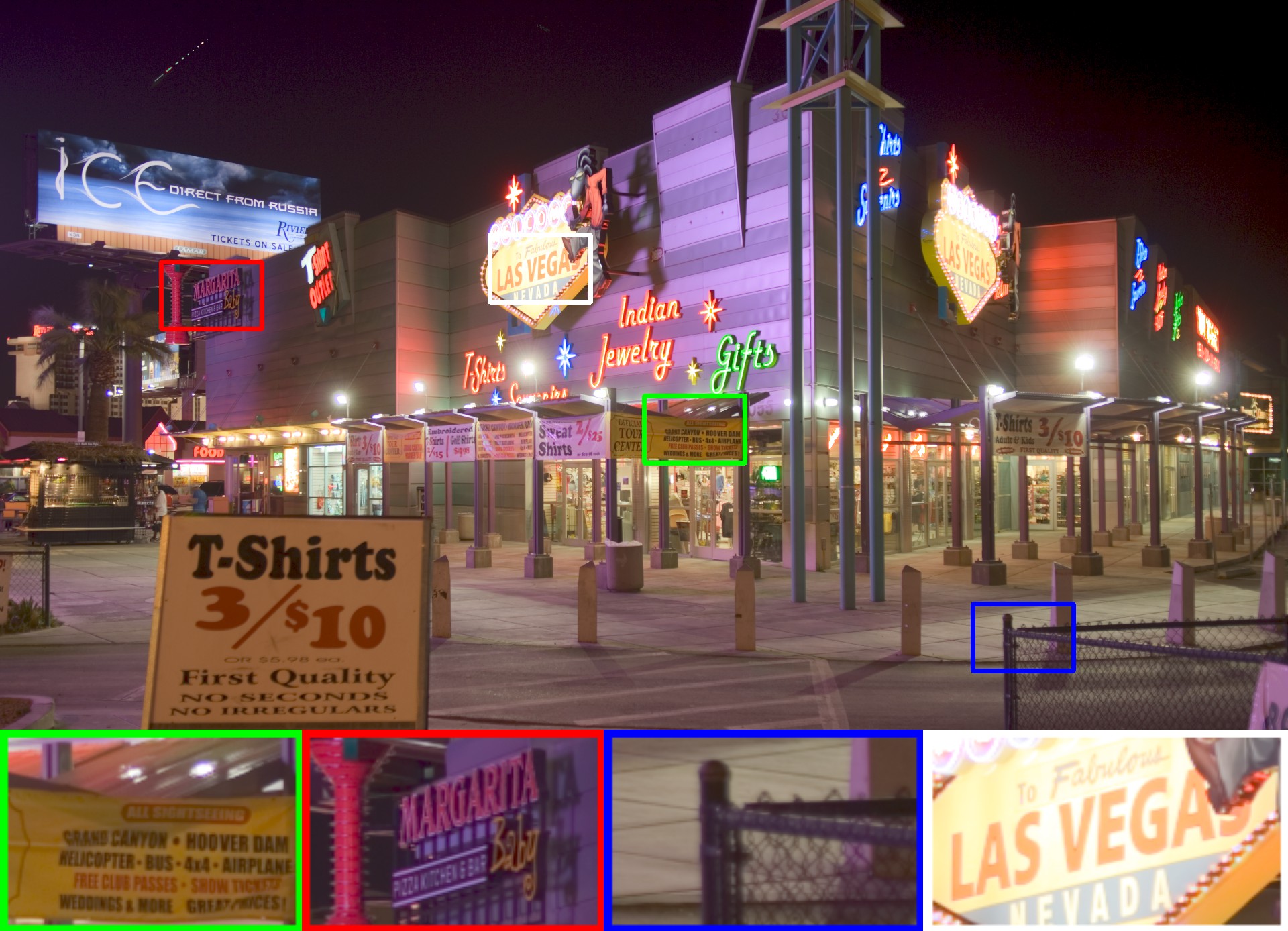} & \includegraphics[width=.25\linewidth]{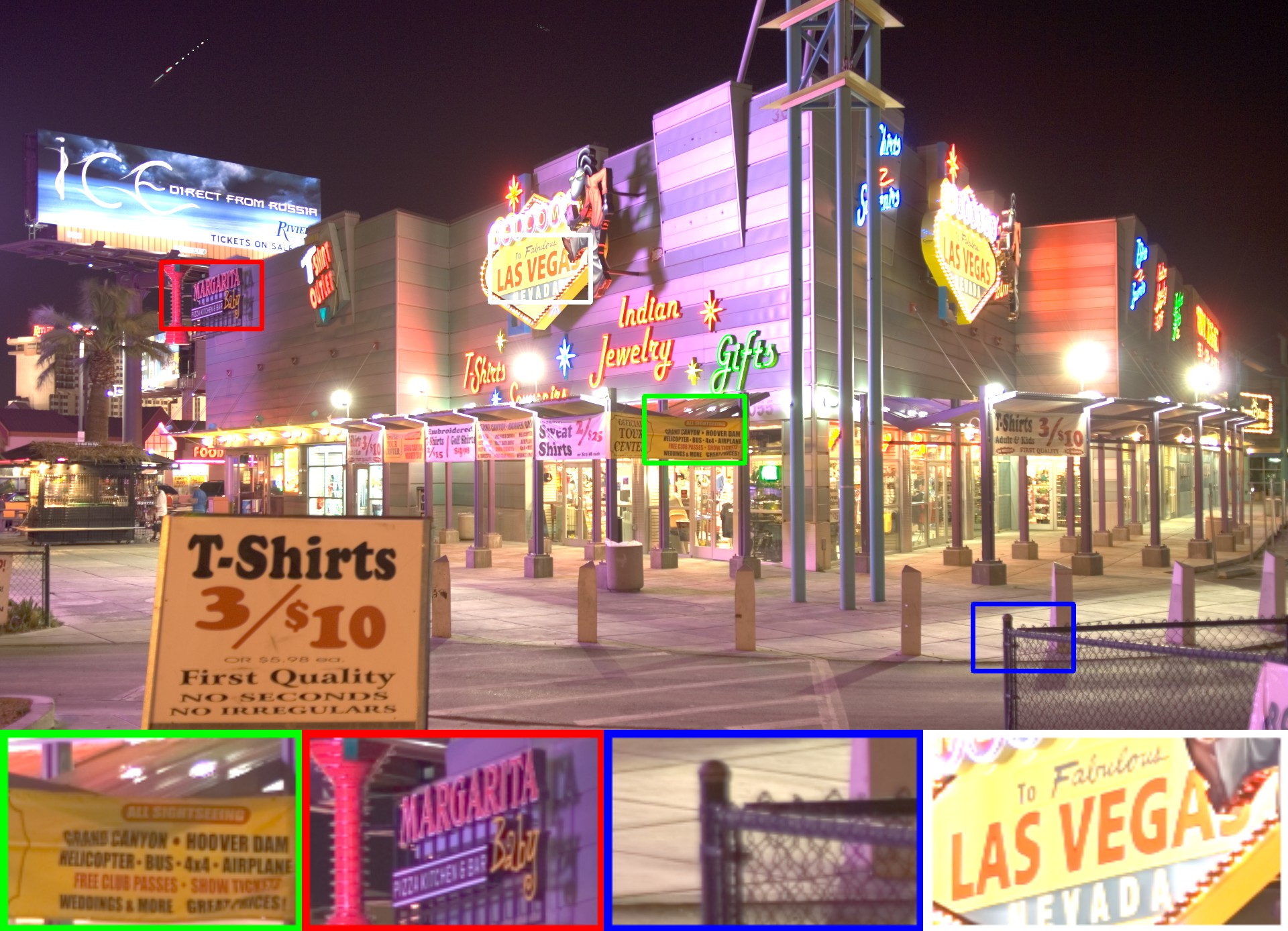} & \includegraphics[width=.25\linewidth]{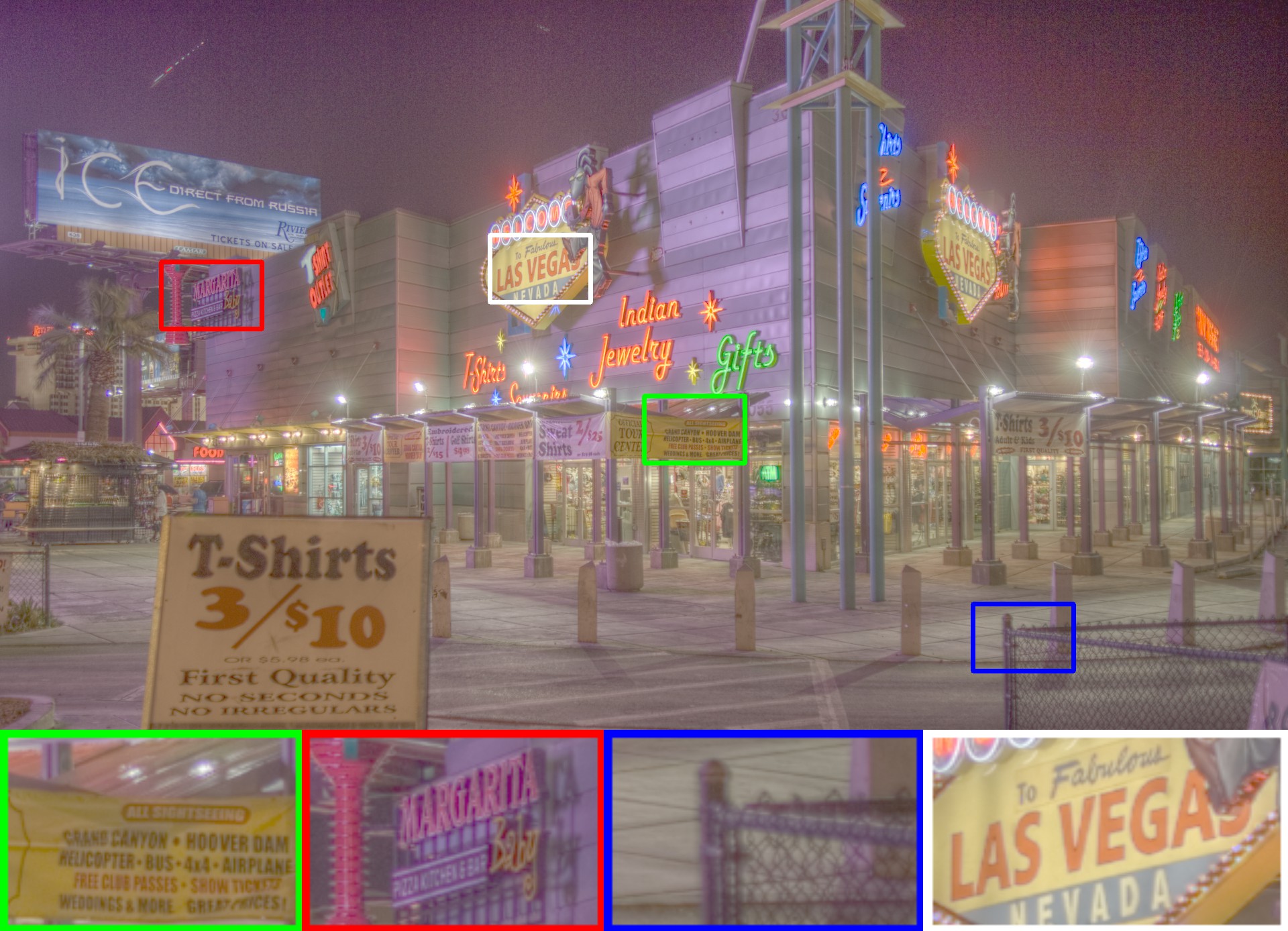} \\
    (a) linear & (b) Mantiuk~\etal~\cite{Mantiuk2008} & (c) Lischinski~\etal~\cite{Lischinski} & (d) Mantiuk~\etal~\cite{Mantiuk:2006:PFC:1166087.1166095} \\
\end{tabular}

\begin{tabular}{cccc}
    \includegraphics[width=.25\linewidth]{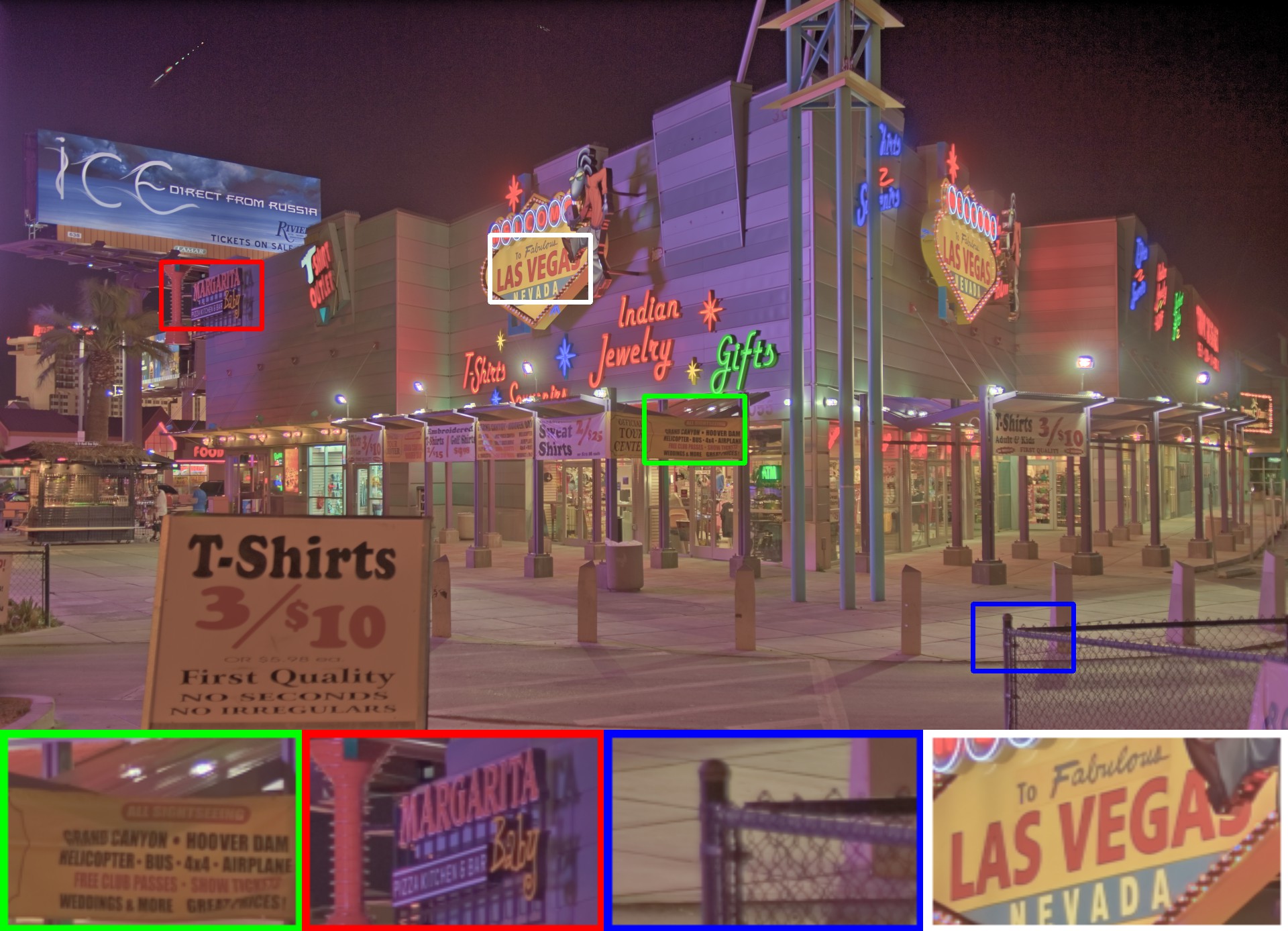} & \includegraphics[width=.25\linewidth]{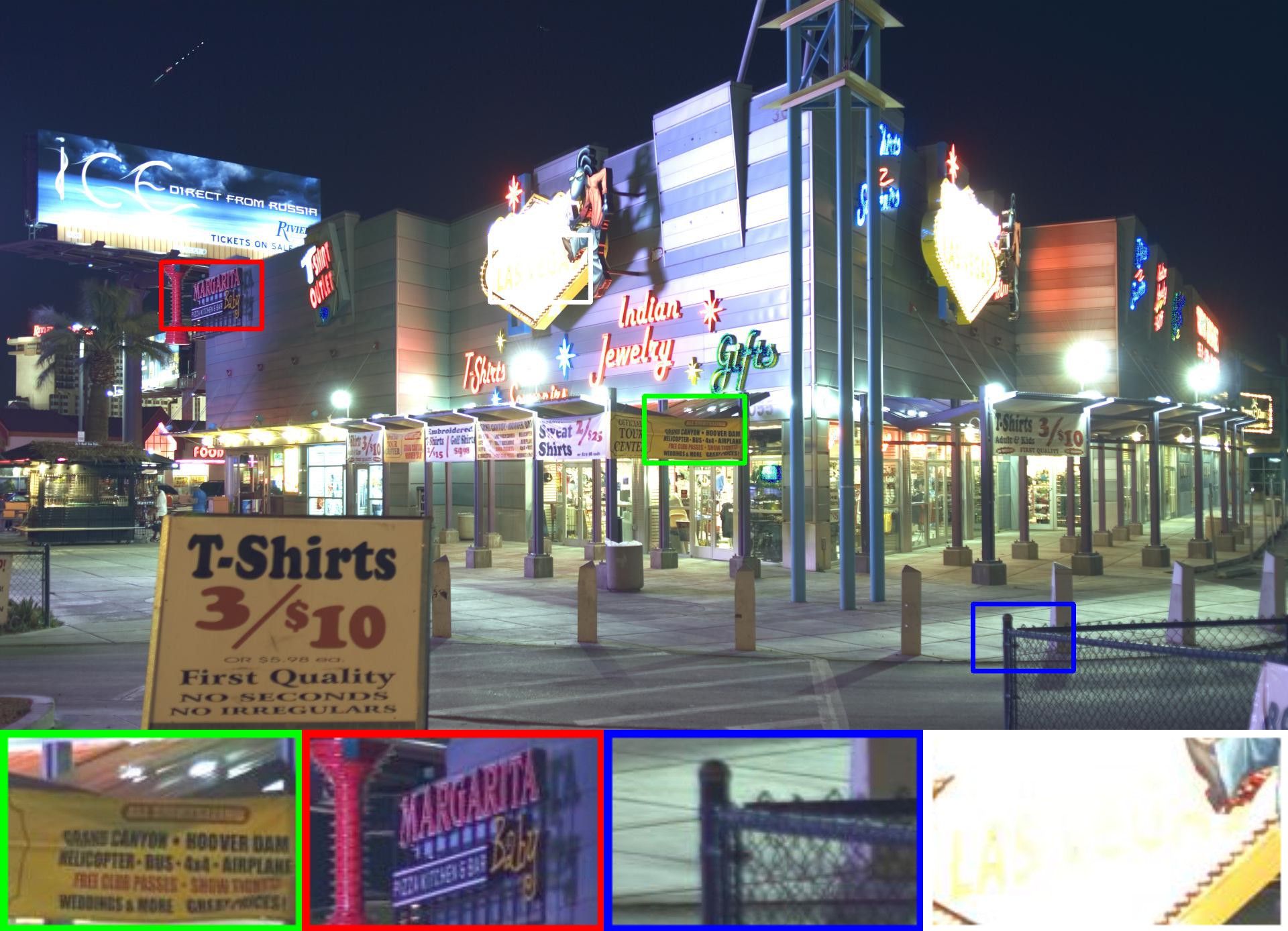} & \includegraphics[width=.25\linewidth]{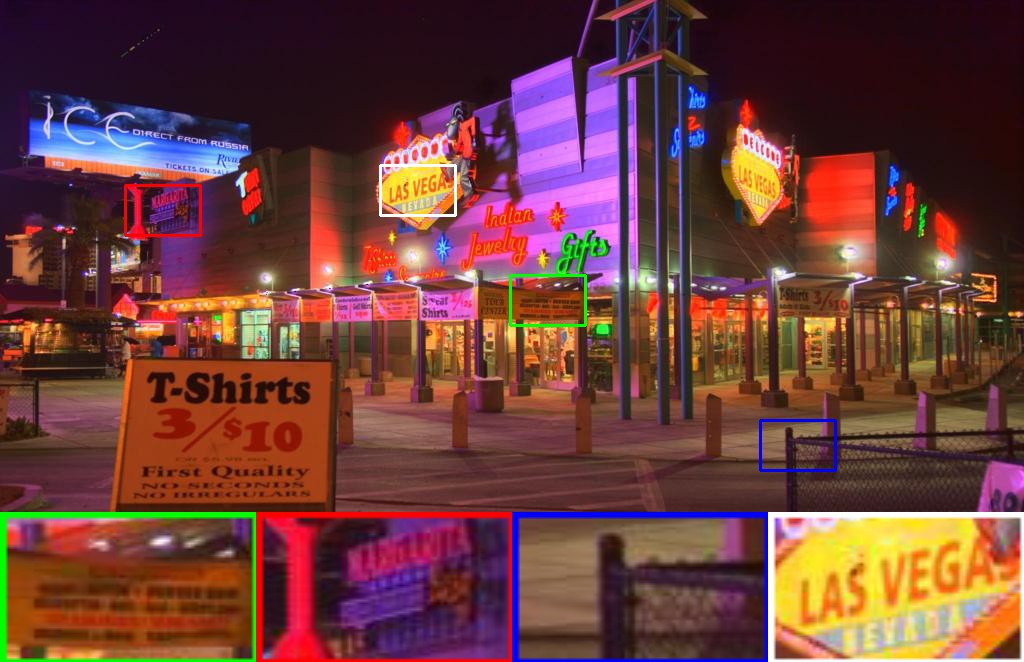} & \includegraphics[width=.25\linewidth]{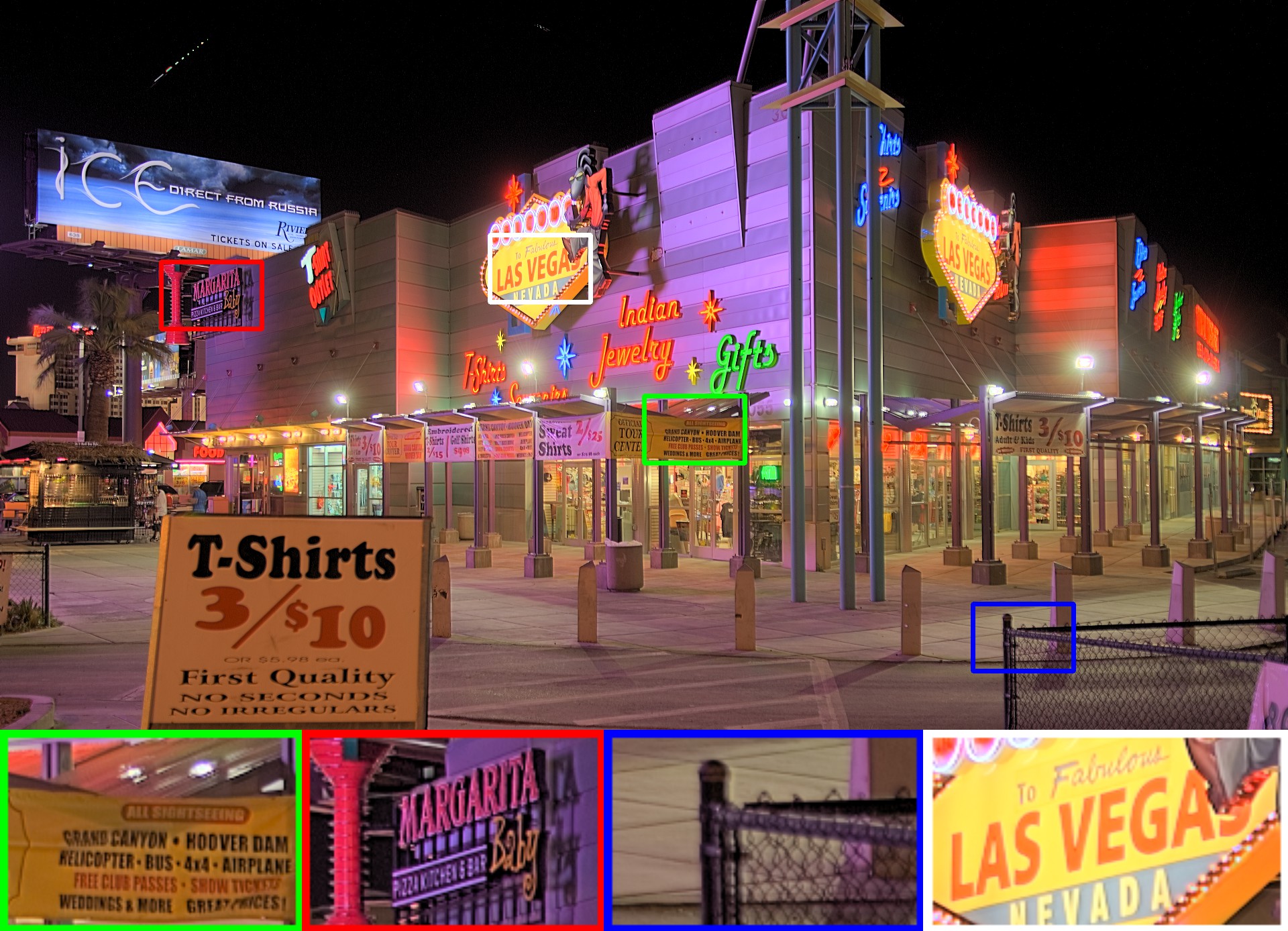} \\
    (e) Liang~\etal~\cite{liang_hybrid_2018} & (f) Yang~\etal~\cite{DRHT} & (g) *Rana~\etal~\cite{rana_deep_2019} & (h) Ours \\
\end{tabular}
%
%\begin{tabular}{cccc}
%    \includegraphics[width=.24\linewidth]{images/result_zoomin_8_figs/cereal_radiance.jpg} & \includegraphics[width=.24\linewidth]{images/result_zoomin_8_figs/cereal_mantiuk08.jpg} & \includegraphics[width=.24\linewidth]{images/result_zoomin_8_figs/cereal_lischinski.jpg} & \includegraphics[width=.24\linewidth]{images/result_zoomin_8_figs/cereal_mantiuk06.jpg} \\
%    (a) linear & (b) Mantiuk~\etal~\cite{Mantiuk2008} & (c) Lischinski~\etal~\cite{Lischinski} & (d) Mantiuk~\etal~\cite{Mantiuk:2006:PFC:1166087.1166095} \\
%\end{tabular}
%

%\begin{tabular}{cccc}
%    \includegraphics[width=.24\linewidth]{images/result_zoomin_8_figs/cereal_L1L0.jpg} & \includegraphics[width=.24\linewidth]{images/result_zoomin_8_figs/cereal_deep.jpg} & \includegraphics[width=.24\linewidth]{images/result_zoomin_8_figs/cereal_TIP.jpg} & \includegraphics[width=.24\linewidth]{images/result_zoomin_8_figs/cereal_ours.jpg} \\
%    (e) Liang~\etal~\cite{liang_hybrid_2018} & (f) Yang~\etal~\cite{DRHT} & (g) *Rana~\etal~\cite{rana_deep_2019} & (h) Ours \\
%\end{tabular}
%
%\captionsetup{width=0.96\linewidth}
\caption{\textbf{Qualitative comparison on tone-mapping methods.} Our results strike a better balance between local detail preserving and global tone compression. (* indicates the results are from their paper. The aspect ratio might be different from others.)}
\label{fig:zoomin_result}
\end{figure*}

We compare our method with three representative conventional methods~\cite{Mantiuk:2006:PFC:1166087.1166095,Mantiuk2008,Lischinski}, one layer decomposition method~\cite{liang_hybrid_2018}, and two deep learning-based methods~\cite{DRHT,rana_deep_2019} on Fairchild's Dataset.
The three conventional methods are implemented by Luminance HDR.
For Liang~\etal~\cite{liang_hybrid_2018}, we use their MATLAB source code with the default parameters.
For Yang~\etal~\cite{DRHT}, we use the pre-trained model provided by the authors.
Unfortunately, the source code of Rana~\etal~\cite{rana_deep_2019} is not available, so we only compare the results reported in their paper.
Fig.~\ref{fig:zoomin_result} shows that our results preserve the detail of input HDR images without halo artifacts or over enhancement.
The color appearance of our results are also visually appealing.
On the contrary, Lischinski~\etal~\cite{Lischinski} often produce over-exposed results, and Liang~\etal~\cite{liang_hybrid_2018} fail to generate natural results.
%
%Note that the result of ours in the figure using zero latent code as input. 
% We then compare our method with other deep learning based methods.
%
% As shown in Fig.~\ref{fig:zoomin_result} (f)
For deep learning-based methods, Yang~\etal~\cite{DRHT} cannot preserve the detail in highlight regions well and generate grid-like artifacts, and Rana~\etal~\cite{rana_deep_2019} show an over-saturation problem. Most importantly, these deep learning-based methods can only support one-to-one mapping, whereas our method is multimodal.
\begin{figure}[htb]
\centering
\renewcommand{\tabcolsep}{1pt} % adjust horizontal space
\renewcommand{\arraystretch}{1} % adjust vertical space
\begin{tabular}{ccc}
    \includegraphics[width=.33\linewidth]{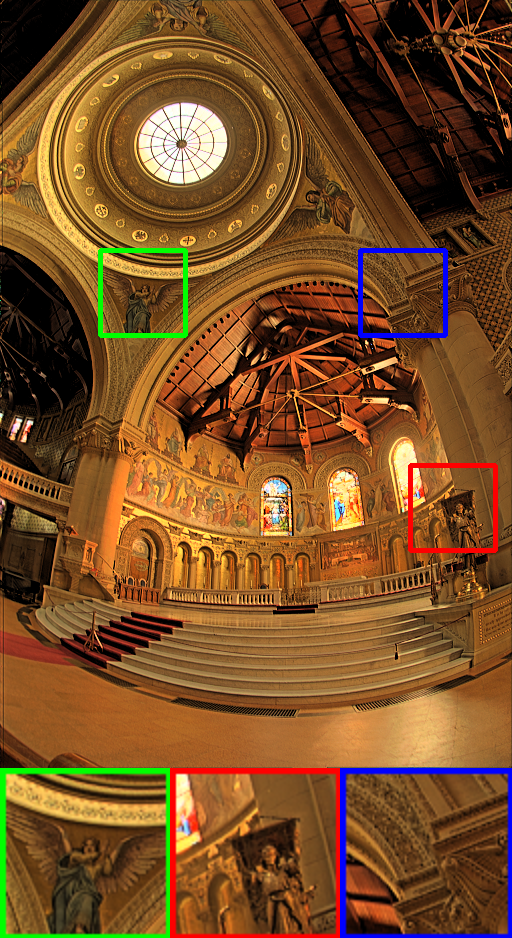} & 
    \includegraphics[width=.33 \linewidth]{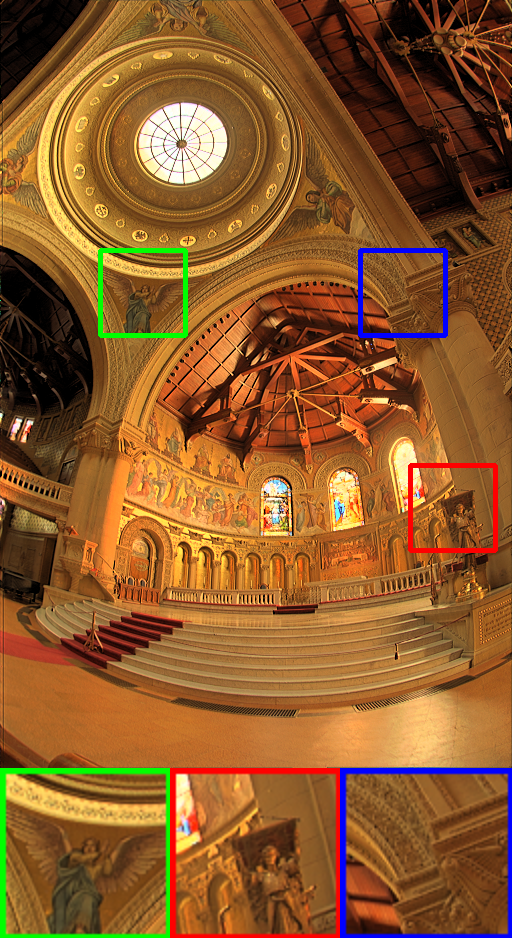} & 
    \includegraphics[width=.33 \linewidth]{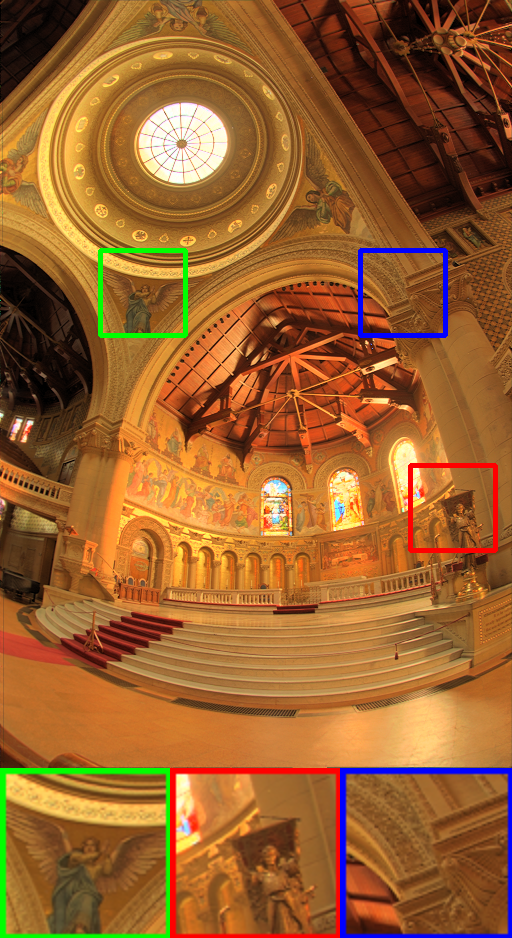}\\
    (a) Large & (b) Medium & (c) Small  \\
\end{tabular}
\caption{\textbf{Different detail strengths in different latent codes.}}
\label{fig:diversity}
\end{figure}

\subsection{Diversity in different latent codes}
As mentioned in Section~\ref{our-pipeline}, our method is multimodal because of the framework of BicycleGAN with various random latent codes during training. Fig.~\ref{fig:teaser} shows the linearity of our results with respect to latent codes. On the other hand, Fig.~\ref{fig:diversity} shows that we can also control the detail strength by adjusting the latent code.
%

%TMQI should be introduced beforehand 
\subsection{Quantitative Comparison}\label{sec:quantitative}
Besides qualitative comparisons, we also perform a quantitative comparison based on TMQI and its sub-metrics.
In Table~\ref{tab:tmqi_score}, we list the methods of top-10 highest TMQI scores on Fairchild’s dataset.
As shown in the table, our method achieves the highest TMQI and naturalness score.
Although Mantiuk~\etal~\cite{Mantiuk:2006:PFC:1166087.1166095} achieve the highest fidelity score among all compared methods, they often over-enhance texture detail and generate unnatural results.
%
% We will address this in~\ref{sec:user_study}

\begin{table}
\centering
\caption{\textbf{Quantitative comparison on tone-mapping methods.} The table shows that our proposed method performs competitively against existing tone-mapping methods.}
\label{tab:tmqi_score}
\begin{tabular}{l|ccc} 
\toprule
% \hline
Methods&TMQI&Fidelity&Naturalness\\ 
\midrule
% \hline
Drago~\etal~\cite{drago}&0.8030&0.7777&0.2289\\
Fattal~\etal~\cite{fattal_gradient_2002}&0.8126&0.8217&0.2220\\
Durand~\etal~\cite{durand_fast_2002}&0.8211&0.7943&0.2993\\
Mai~\etal~\cite{mai}&0.8184&0.8077&0.2554\\
Reinhard~\etal~\cite{reinhard_photographic_2002}&0.8282&0.7964&0.3281\\
Mantiuk 08~\etal~\cite{Mantiuk2008}&0.8447&0.8336&0.3496\\
Lischinski~\etal~\cite{Lischinski}&0.8564&0.8381&0.4110\\
Mantiuk 06~\etal~\cite{Mantiuk:2006:PFC:1166087.1166095}&0.8574&\pmb{0.8797}&0.3412\\
Liang~\etal~\cite{liang_hybrid_2018}&0.8650&0.8074&0.5024\\
Yang~\etal~\cite{DRHT}&0.8465&0.7564&0.4793\\
Ours&\pmb{0.8652}&0.8064&\pmb{0.5096}\\
\bottomrule
% \hline
\end{tabular}
\end{table}

\subsection{User Study}
\label{sec:user_study}
To further verify the performance of our method, we conduct a user study on Fairchild's dataset~\cite{Fairchild}.
Totally, 40 subjects are involved in this test.
We choose the methods of top-4 highest TMQI scores on Fairchild’s dataset for comparison.
The subjects are asked to select a preferable image for each pair comparison of the dataset.
Then, the results are evaluated by calculating the probability of being selected by the subjects.
Fig.~\ref{fig:user_study} shows that our results achieve higher probability than the compared methods.

\section{Conclusions}
\label{sec:conclusion}

We have presented a novel deep learning-based tone-mapping method. The proposed method performs favorably against existing conventional methods and learning-based methods quantitatively and qualitatively.
We have also provided a user study to make the experimental results more convincing. Moreover, the proposed method can produce a variety of expert-level tone-mapped results by adjusting the latent code.

As for future work, making the latent code more explainable and easier to adjust could be a possible direction.

%Despite the complex tuning process of the conventional methods and the uncontrollablity of the previous deep learning based methods, the proposed method which uses an improved cVAE-GAN to perform tone mapping operator is steerable through the latent code manipulation. Different styles of tone mapping could be performed easily by the proposed method. The results achieves the best visual quality both in objective and subjective comparison. 

%All other methods are just joking.

\clearpage

\bibliographystyle{IEEEtran}
\bibliography{egbib}
%\begin{thebibliography}{1}


%\end{thebibliography}

\end{document}